\documentclass{article}
\usepackage{iclr2027_conference,times}
\iclrfinalcopy 

\usepackage[T1]{fontenc}
\usepackage[utf8]{inputenc}
\usepackage{amsmath,amssymb,amsthm,mathtools}
\usepackage{booktabs}
\usepackage{array,tabularx,multirow,longtable}
\usepackage{graphicx}
\usepackage{wrapfig}
\usepackage{flafter}
\usepackage{needspace}
\usepackage{float}
\usepackage{algorithm}
\usepackage{algorithmic}
\usepackage{microtype}
\usepackage{listings}
\usepackage{xcolor}
\usepackage{tikz}
\usetikzlibrary{arrows.meta,positioning,fit,calc,backgrounds}
\usepackage{tcolorbox}
\tcbuselibrary{listings,breakable,skins}
\usepackage{placeins}
\usepackage[colorlinks=true,linkcolor=blue,citecolor=blue,urlcolor=blue]{hyperref}
\usepackage{url}

\definecolor{figBlue}{HTML}{3F6F9F}
\definecolor{figBlueFill}{HTML}{EDF3F8}
\definecolor{figGreen}{HTML}{3D7A68}
\definecolor{figGreenFill}{HTML}{EDF6F2}
\definecolor{figOrange}{HTML}{A9652A}
\definecolor{figOrangeFill}{HTML}{FAF2E8}
\definecolor{figRed}{HTML}{955050}
\definecolor{figRedFill}{HTML}{F8EEEE}
\definecolor{figGray}{HTML}{59636D}
\definecolor{figGrayFill}{HTML}{F2F4F5}
\definecolor{figPurple}{HTML}{6335A8}
\definecolor{figPurpleFill}{HTML}{F2E9FC}

\newtheorem{theorem}{Theorem}
\newtheorem{proposition}{Proposition}
\newtheorem{corollary}{Corollary}
\newtheorem{lemma}{Lemma}
\theoremstyle{definition}
\theoremstyle{plain}
\newtheorem{assumption}{Assumption}

\newcommand{\F}{\mathcal F}
\newcommand{\Fpre}{\mathcal F^{\mathrm{pre}}}
\newcommand{\E}{\mathbb E}
\newcommand{\Prb}{\mathbb P}
\newcommand{\ind}{\mathbf 1}
\newcommand{\Thetac}{\widetilde\Theta}
\newcommand{\FER}{\operatorname{FER}}
\newcommand{\TER}{\operatorname{TER}}
\newcommand{\clip}{\operatorname{clip}}
\newcommand{\TRUST}{TRUST}

\AtBeginDocument{%
\newtcblisting[use counter*=lstlisting,list inside=lol,list type=lstlisting]{caseListing}[1]{%
  enhanced,breakable,listing only,listing engine=listings,
  listing file={\jobname-case.listing},
  width=\linewidth,colback=white,colbacktitle=figGrayFill,
  colframe=figGray!55,coltitle=black,
  boxrule=0.45pt,titlerule=0.35pt,arc=1.2pt,outer arc=1.2pt,
  boxsep=0pt,left=7pt,right=7pt,top=6pt,bottom=6pt,
  lefttitle=7pt,righttitle=7pt,toptitle=5pt,bottomtitle=5pt,
  fonttitle=\normalfont\normalsize,
  title={\textbf{Listing~\thetcbcounter}\enspace#1},list text={#1},
  title after break={\textbf{Listing~\thetcbcounter}\enspace#1\ \textit{(continued)}},
  before skip=9pt plus 2pt minus 1pt,after skip=8pt plus 2pt minus 1pt,
  lines before break=4,pad at break=3pt,
  listing options={basicstyle=\ttfamily\normalsize,
    breaklines=true,breakatwhitespace=false,columns=fullflexible,
    keepspaces=true,showstringspaces=false,numbers=none,
    frame=none,backgroundcolor={},xleftmargin=0pt,xrightmargin=0pt,
    aboveskip=0pt,belowskip=0pt,
    literate={≤}{{$\leq$}}1 {—}{{\textemdash}}1 { }{{ }}1}}
}
\hypersetup{pdftitle={Trustworthy Recursive Self-Improvement with Statistical Guarantees},pdfauthor={Zihe Gao, Zhengling Qi, Yifan Cui}}

\title{\raggedright\exhyphenpenalty=0\relax Trustworthy Recursive Self-Improvement with Statistical Guarantees}
\author{%
\makebox[\dimexpr\textwidth-2\tabcolsep\relax][c]{%
  \textbf{Zihe Gao}\textsuperscript{1}\qquad
  \textbf{Zhengling Qi}\textsuperscript{2}\qquad
  \textbf{Yifan Cui}\textsuperscript{1}}\\[0.4em]
\makebox[\dimexpr\textwidth-2\tabcolsep\relax][c]{%
  \normalfont\textsuperscript{1}Zhejiang University}\\
\makebox[\dimexpr\textwidth-2\tabcolsep\relax][c]{%
  \normalfont\textsuperscript{2}George Washington University}%
}

\begin{document}
\maketitle
\fancyhead{}
\renewcommand{\headrulewidth}{0.4pt}

\begin{abstract}
Recursive self-improvement offers a promising path toward AI agents that improve through experience by refining their own harnesses without retraining their underlying large language models (LLMs). Making this self-improvement process reliable, however, requires more than generating new updates: agents must decide which changes are worth keeping. A greedy gate that accepts updates based solely on higher validation scores can mistake an observed gain for genuine progress and lead to performance degeneration, whereas overly conservative statistical gates may reject beneficial updates and hinder performance improvement. We introduce Trustworthy Recursive Updates with Statistical Testing (TRUST), a framework that adapts an online multiple hypothesis testing procedure to control the rate of false evolution  across rounds, with an LLM analyzing historical context to help determine how much evidence is required to update the incumbent. Under interpretable assumptions, we prove that TRUST limits the proportion of accepted updates that result in unacceptable performance loss. We further derive a lower bound on the final performance gain. Across the benchmarks HotpotQA, DROP, and APPS with the models DeepSeek-V4-Flash and Qwen-3.8-Flash, TRUST reduces the average ratio of accepted inferior updates from the greedy gate's 45.59\% to 2.78\%, while achieving higher final scores than the state-of-the-art statistical gate baselines for recursive self-improvement. These findings demonstrate that TRUST strikes a balance between statistical reliability and performance gain in recursive self-improvement.
\end{abstract}

\section{Introduction}
Recursive self-improvement (RSI) has attracted growing interest for its potential to enhance agents' capabilities without retraining their underlying language models \citep{ren2026selfimprovements,chen2026recursivesurvey}. In a self-improvement loop, agents modify their own prompts, memory, tools, and workflows to improve task performance \citep{madaan2023selfrefine,shinn2023reflexion,wang2023voyager}. This loop can be organized into three components: a solver that executes tasks, a proposer that generates candidate updates, and a gate that decides whether to accept them \citep{wang2026diving,ren2026selfimprovements}.

\begin{wrapfigure}{r}{.32\textwidth}
\centering
\includegraphics[width=\linewidth]{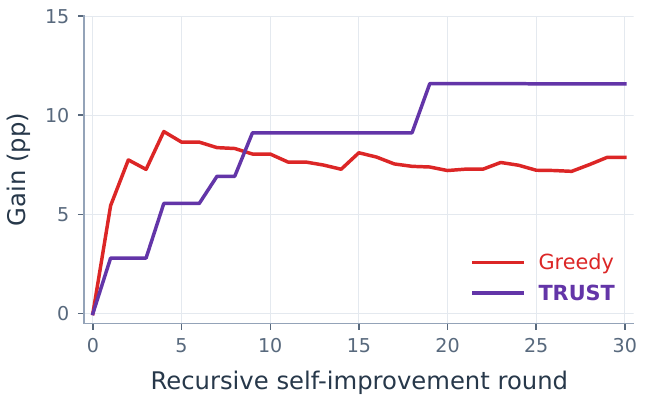}
\caption{Gain trajectories of TRUST and Greedy on HotpotQA with DeepSeek-V4-Flash. Details are in Section~\ref{sec:main-gate-comparison}.}
\label{fig:intro-hotpot-trajectories}
\end{wrapfigure}

\begin{figure}[t]
\centering
\includegraphics[width=\linewidth]{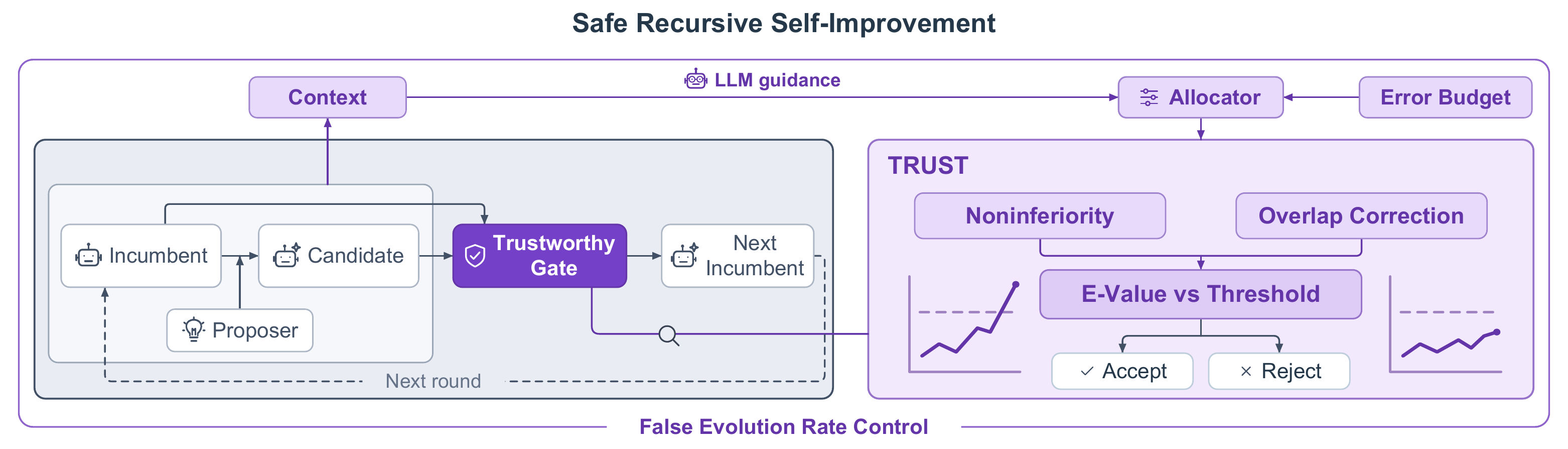}
\caption{Recursive self-improvement with our trustworthy gate TRUST. \TRUST{} combines noninferiority, overlap correction, e-value testing and error budget allocation with FER control across rounds.}
\label{fig:trust-framework-overview}
\end{figure}

Existing work has explored mechanisms for generating better candidate updates, while less attention has been paid to deciding when to accept them. A common acceptance approach is to use a greedy gate, which accepts a candidate whenever its score on a fixed validation set exceeds the incumbent's \citep{pace2026}. Figure~\ref{fig:intro-hotpot-trajectories} illustrates how early gain under greedy acceptance can be lost in subsequent rounds. Noise from finite samples and overfitting from reusing validation data can both make inferior updates appear beneficial and pass a greedy gate \citep{dwork2015reusable,rogers2016maxinformation}. Accepting a harmful update can reduce the agent's current performance. The resulting loss may persist through subsequent rounds, ultimately limiting the gain for self-improvement.

Recent work has begun to focus on the acceptance gate \citep{wu2025sgm,pace2026,sengupta2026sea}. RSEA uses an extra validation set separate from the proposal generation set to select candidate updates \citep{nguyen2026rsea}. PACE provides a safety guarantee for each candidate update through an e-process test in a single round instead of the entire recursive self-improvement process \citep{pace2026}. However, controlling the probability of any false acceptance across rounds can make statistical gates conservative and hinder beneficial updates \citep{wu2025sgm,sengupta2026sea}. A key challenge, therefore, remains: retaining opportunities for useful improvement while controlling acceptance errors across rounds under limited validation resources.

We introduce Trustworthy Recursive Updates with Statistical Testing (TRUST), a general statistical acceptance framework.  Motivated by false discovery rate (FDR) control \citep{benjamini1995controlling,xu2023online,zhang2025egai}, we define the false evolution rate (FER) as a safety criterion for recursive self-improvement. FER measures the expected proportion of accepted updates that cause an unacceptable performance loss, thereby quantifying the gate's false acceptance risk. To control FER, we formulate the sequential acceptance decisions as an online multiple testing problem. We also use an LLM to analyze historical context and help the error budget allocator assign less restrictive testing levels to promising candidates. We further draw on noninferiority testing to improve statistical power by allowing a possible loss tolerance \citep{blackwelder1982proving,tunes2009methods,schumi2011looking}, and use overlap correction to account for bias from reusing validation samples across rounds \citep{dwork2015reusable,rogers2016maxinformation}. Figure~\ref{fig:trust-framework-overview} summarizes how TRUST makes recursive self-improvement trustworthy.

\noindent\textbf{Our contributions.} We make the following three main contributions: (a) Our framework builds on online multiple hypothesis testing with interpretable parameters, including the error budget, loss tolerance, and overlap correction strength. It naturally incorporates LLM guidance based on diverse forms of historical context. (b) We establish statistical guarantees under explicit assumptions, including FER control, power consistency for improving candidates, and a lower bound on deployment gain. These results connect hypothesis testing criteria to performance gain in recursive self-improvement. (c) We evaluate TRUST in a closed-loop self-improvement harness  across HotpotQA, DROP, and APPS with DeepSeek-V4-Flash and Qwen-3.8-Flash. Averaged across these settings, TRUST achieves the highest final gain of 24.84~pp. Compared with the greedy gate, TRUST reduces the mean FER$_0$ from 45.59\% to 2.78\% and the mean maximum drawdown from 4.963~pp to 0.047~pp, demonstrating high performance gains with few erroneous acceptances and small drawdowns.

\section{Related Work}
\label{sec:related-work}

\paragraph{Recursive self-improvement and agent harness.}
An agent harness includes tools, context management, and execution control \citep{shi2026programming}. Recursive self-improvement agents use feedback to improve various harness components. Promptbreeder evolves task prompts \citep{fernando2024promptbreeder}. Reflexion, ExpeL, and Voyager convert experience into reflections, insights, and reusable skills, respectively  \citep{shinn2023reflexion,zhao2024expel,wang2023voyager}. CoEvo-Mem improves memory retrieval \citep{ye2026coevomem}. AFlow and Flow refine agentic workflows \citep{zhang2025aflow,niu2025flow}. AgentSquare defines a modular design space with planning, reasoning, tool use, and memory components \citep{shang2025agentsquare}. These works focus on proposing effective candidates, while we study whether a new candidate should be accepted to make recursive self-improvement trustworthy. Additionally, we design a harness with diverse editable components in our experiments to evaluate how different gates affect performance and risk.

\paragraph{Hypothesis testing methods.} Multiple hypothesis testing procedures are primarily based on two error criteria: the familywise error rate (FWER) and the FDR \citep{benjamini1995controlling,wang2022fdr,robertson2023online}. FWER control is generally more conservative than FDR control. Many popular online FDR methods with p-values rely on independence across tests \citep{ramdas2018saffron,robertson2023online}. This independence can be difficult to satisfy when agents evolve based on past feedback. The e-value is an alternative test statistic with a betting interpretation \citep{shafer2021testing}. The e-BH and e-LOND procedures allow arbitrary dependence among valid e-values while controlling FDR \citep{wang2022fdr,xu2023online}. Drawing on e-GAI \citep{zhang2025egai} and contextual online testing \citep{chen2020contextual}, TRUST constructs an e-value and allocates the testing levels with context. In RSI, SGM uses statistical tests to decide updates to control FWER across rounds \citep{wu2025sgm}, while PACE uses an e-process to control the probability of false acceptance in a single round \citep{pace2026}. TRUST provides FER control across rounds and achieves higher final scores than these statistical baselines in our experiments.

\section{Problem Formulation}
\label{sec:setup}

Let $\mathcal D$ denote the distribution of tasks and let $u(x,s)\in[0,1]$ denote the score of output $s$ on task $x$. For a solver $\theta$, let $S_\theta(x,\Omega)$ denote the solver output on task $x$ under execution randomness $\Omega$. Define the population score of $\theta$ as
\begin{equation}
J(\theta)
=
\E_{X\sim\mathcal D,\Omega}
\left[
u\bigl(X,S_\theta(X,\Omega)\bigr)
\right].
\label{eq:population-utility}
\end{equation}
We consider a self-improvement process that starts with an initial incumbent solver $\Theta_0$ and initial information $\mathcal{F}_0$. It recursively generates and evaluates candidates over $T$ rounds. In round $t$, let $Q(\cdot\mid\mathcal{F}_{t-1})$ denote the conditional distribution of the candidate given the completed history $\mathcal{F}_{t-1}$ from the fixed proposer. The proposer generates
$\widetilde{\Theta}_t
\sim
Q\!\left(\cdot\mid\mathcal{F}_{t-1}\right)$,
which is evaluated by the gate $\mathcal G$ against the previous incumbent
$\Theta_{t-1}$. Then the gate $\mathcal{G}$ produces $A_t\in\{0,1\}$ and updates the incumbent  according to
\begin{equation*}
\Theta_t
=
\begin{cases}
\widetilde{\Theta}_t, & A_t=1,\\
\Theta_{t-1}, & A_t=0.
\end{cases}
\end{equation*}
Different gates may induce different acceptance decisions, thereby changing the subsequent self-improvement trajectories and the candidates proposed in later rounds. These changes can affect the final gain, which is defined as
\begin{equation}
G(T)=\mathbb{E}\!\left[J(\Theta_T)-J(\Theta_0)\right].
\label{eq:gain}
\end{equation}

We next define the core safety criterion. The population score difference between the selected candidate and the incumbent is $\Delta_t = J\!\left(\widetilde{\Theta}_t\right) - J\!\left(\Theta_{t-1}\right).$ For a given tolerance level $\varepsilon\in[0,1)$, let $N_t^{\varepsilon}=\ind\{\Delta_t\le-\varepsilon\}$,
$R_T=\sum_{t\le T}A_t$, and $F_T^\varepsilon=\sum_{t\le T}N_t^{\varepsilon}A_t$. Define the false evolution rate (FER) at tolerance $\varepsilon$ as
\begin{equation}
\FER_\varepsilon(T)
=\E\left[\frac{F_T^\varepsilon}{R_T\vee1}\right].
\label{eq:fer-definition}
\end{equation}

Here, FER is the expected proportion of unacceptably inferior updates among accepted updates, and $\varepsilon$ specifies the tolerated loss. For \(\varepsilon>0\), the difference \(\Delta_t\) is divided into three regions: updates with \(\Delta_t\le -\varepsilon\) are unacceptable, those with \(-\varepsilon<\Delta_t\le 0\) are tolerated but not improving, and those with \(\Delta_t>0\) are improving. Our goal is to ensure that $\operatorname{FER}_{\varepsilon}(T)\le q$ for a prespecified FER budget $q$, thereby providing a statistical safety guarantee for the entire self-improvement process. This helps avoid unacceptable updates, retain beneficial proposals, and may achieve a favorable final gain.

Let $\mathcal V_N=(Z_1,\ldots,Z_N)\sim\mathcal D^N$ be a fixed  validation pool of size $N$. Candidate generation uses information from the completed history. In round $t$, the sampling plan is to draw $n\le N$ tasks uniformly without replacement from $\mathcal V_N$ to evaluate the current candidate. All sampled tasks are then returned to $\mathcal V_N$ so that the next round starts with the full validation pool. Let $K_{t,i}$ denote the pool index selected at the $i$th validation draw, and let
$X_{t,i}=Z_{K_{t,i}}$ be the corresponding task. On task $X_{t,i}$, the candidate and incumbent produce the observed score difference
\begin{equation}
D_{t,i}
={}
u\bigl(
X_{t,i},
S_{\widetilde{\Theta}_t}(X_{t,i},\Omega^c_{t,i})
\bigr)
-
u\bigl(
X_{t,i},
S_{\Theta_{t-1}}(X_{t,i},\Omega^0_{t,i})
\bigr)
\in[-1,1],
\label{eq:paired-difference}
\end{equation}
where $\Omega^c_{t,i}$ and $\Omega^0_{t,i}$ denote the execution
randomness of the candidate and incumbent, respectively.

\begin{figure}[H]
\centering
\resizebox{\linewidth}{!}{
\begin{tikzpicture}[
  x=1cm,y=0.7cm,
  font=\sffamily\fontsize{8}{10}\selectfont,
  text=black!85,
  stage/.style={align=center,font=\sffamily\fontsize{8}{10}\selectfont},
  history/.style={font=\fontsize{11}{13}\selectfont},
  detail/.style={align=center,anchor=north,text width=2.9cm},
  timepoint/.style={circle,fill=figBlue,inner sep=2.1pt}
]
\path[use as bounding box] (0,.9) rectangle (16.85,4.15);
\fill[figBlueFill,rounded corners=3pt] (0,.9) rectangle (16.85,4.15);
\draw[-{Stealth[length=2.2mm]},figBlue!60,line width=.7pt]
  (.4,2.75) -- (16.25,2.75);

\foreach \x in {1.35,4.65,8.0,11.7,15.15}
  \node[timepoint] at (\x,2.75) {};
\node[stage] at (1.35,3.75) {Before round $t$};
\node[stage] at (4.65,3.75) {Before validation};
\node[stage] at (8.0,3.75) {First sample observed};
\node[stage] at (11.7,3.75) {$n$ samples observed};
\node[stage] at (15.15,3.75) {After validation};
\node[history] at (1.35,3.20) {$\F_{t-1}$};
\node[history] at (4.65,3.20) {$\Fpre_t=\mathcal H_{t,0}$};
\node[history] at (8.0,3.20) {$\mathcal H_{t,1}$};
\node[history] at (11.7,3.20) {$\mathcal H_{t,n}$};
\node[history] at (15.15,3.20) {$\F_t$};
\node[fill=figBlueFill,inner xsep=5pt,history] at (9.85,2.75) {$\cdots$};

\node[detail,text width=2.55cm] at (1.35,2.40)
  {Completed history};
\node[detail,text width=3.1cm] at (4.65,2.40)
  {Selected candidate $\Thetac_t$\\Testing level $\alpha_t$};
\node[detail,text width=3.0cm] at (8.0,2.40)
  {First sample\\\mbox{Paired outcomes: $D_{t,1}$}};
\node[detail,text width=3.7cm] at (11.7,2.40)
  {$n$ samples\\\mbox{Paired outcomes: $\{D_{t,i}\}_{i=1}^{n}$}};
\node[detail,text width=3.1cm] at (15.15,2.40)
  {\mbox{Acceptance decision $A_t$}\\\mbox{Updated incumbent $\Theta_t$}};

\end{tikzpicture}}
\caption{Information flow in a self-improvement round: $\F_{t-1}\subseteq\Fpre_t=\mathcal H_{t,0}\subseteq\mathcal H_{t,1}\subseteq\cdots\subseteq\mathcal H_{t,n}\subseteq\F_t\subseteq\Fpre_{t+1}$. The relation $\subseteq$ denotes inclusion of information.}
\label{fig:information-timeline}
\end{figure}
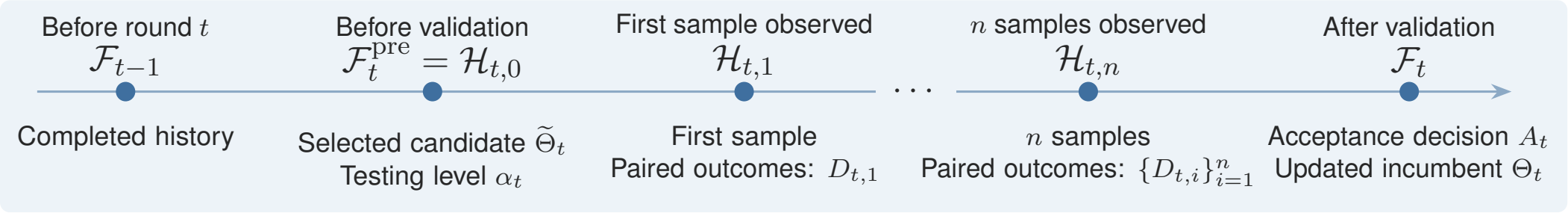

\paragraph{Information flow.}
Our framework uses filtrations to describe how information accumulates within and across rounds \citep{ramdas2023game}. For clarity, we briefly explain this information flow. The initial information $\mathcal F_0$ is independent of the validation pool. Let $\F_{t-1}$ denote the completed history before round $t$. Before validation,
$\Fpre_t$ additionally fixes the selected candidate $\widetilde{\Theta}_t$ and
testing level $\alpha_t$ in Section~\ref{sec:TRUST}. For $i=1,\ldots,n$, let $\mathcal H_{t,i}$ augment
$\Fpre_t$ with the first $i$ validation samples and their paired outcomes,
with $\mathcal H_{t,0}=\Fpre_t$. After validation, $\F_t$ contains the decision $A_t$ and updated incumbent $\Theta_t$. We illustrate this timeline in Figure~\ref{fig:information-timeline}.

\section{Trustworthy Recursive Updates with Statistical Testing}
\label{sec:TRUST}

The greedy gate accepts a candidate whenever the empirical score difference exceeds zero, making the acceptance decision sensitive to sampling and execution randomness. We formulate candidate acceptance as a hypothesis testing problem. In round $t$, we test
\begin{equation}
H_t^0:\Delta_t\le-\varepsilon
,\qquad
H_t^1:\Delta_t>-\varepsilon.
\label{eq:noninferiority-test}
\end{equation}
If the null hypothesis $H_t^0$ is rejected, the selected candidate is accepted as the new incumbent. Otherwise, the previous incumbent is retained.

In online FDR control, testing levels are allocated across multiple hypotheses under an overall error budget \citep{chen2020contextual,zhang2025egai}. TRUST adapts this framework to the sequence of candidate acceptance tests defined in \eqref{eq:noninferiority-test}, using $q$ as the total FER budget. In round $t$, the testing level $\alpha_t$ determines the acceptance threshold, while the test statistic $M_{t,m}^{(\beta)}$ measures the evidence accumulated from the first $m$ validation observations. For $\alpha_t>0$, TRUST accepts the candidate at the first $m\le n$ when $M_{t,m}^{(\beta)}\ge1/\alpha_t$, following the standard testing rule for sequential e-value testing \citep{ramdas2023game}. If the threshold is not reached within $n$ observations, the incumbent is retained. When $\alpha_t=0$, validation is skipped and the incumbent is retained. We next describe how $\alpha_t$ is allocated and how $M_{t,m}^{(\beta)}$ is constructed.

\subsection{Testing Level Allocation with Context}
\label{sec:contextual-allocation}
We now describe how TRUST allocates the testing level $\alpha_t$ in round $t$ under the total FER budget $q$. Let $B_t$ denote the remaining FER budget before round $t$, with $B_1=q\in(0,1)$, and let $\rho_t\in(0,1)$ denote the spending fraction of the remaining FER budget in round $t$.

Adapting the e-LORD allocation rule within the e-GAI framework \citep{zhang2025egai}, we combine the adjusted spending fraction $\rho_t$ with the remaining budget and the number of accepted updates to determine the testing level
\begin{equation}
\alpha_t
=
\min\{\alpha_{\max},\rho_tB_t(R_{t-1}+1)\},
\label{eq:testing-level}
\end{equation}
where $\alpha_{\max}\in(0,1)$.
The resulting charge $c_t$ for the FER budget and the remaining budget $B_{t+1}$ are
\begin{equation}
c_t=\frac{\alpha_t}{R_{t-1}+1},
\qquad
B_{t+1}=B_t-c_t.
\label{eq:budget-update}
\end{equation}
The charge satisfies $0\le c_t\le\rho_tB_t\le B_t$, keeping the balance nonnegative.

We next describe how to determine $\rho_t$. Building on the contextual error budget allocation in online multiple hypothesis testing \citep{chen2020contextual}, we make use of the context $C_t$ in round $t$ and use an LLM to analyze it. This context only contains information available before the current round's validation, including the candidate update and feedback from previous validation rounds. Following the evaluation structure of G-Eval \citep{liu2023geval}, the LLM uses explicit criteria to assess the candidate's potential benefits and risks. Let $\Psi$ denote the mapping from context to LLM multipliers, with $w_t=\Psi(C_t)\in[0,\infty)$. The default spending fraction is set to $\rho_t^0=1/(T-t+1)$, corresponding to an equal share of the remaining FER budget. The LLM multiplier adjusts the spending fraction within the range $[\rho_{\min},\rho_{\max}]$:
\begin{equation}
\rho_t
=
\clip(w_t\rho_t^0,\rho_{\min},\rho_{\max}),
\label{eq:spending-fraction}
\end{equation}
where $0<\rho_{\min}<\rho_{\max}<1$ and $\operatorname{clip}(x,a,b)=\min\{b,\max\{a,x\}\}$. All allocation quantities are determined before validation and included in the information $\Fpre_t$ in Figure~\ref{fig:information-timeline}.

\subsection{E-Value Construction with Overlap Correction}
\label{sec:evidence}
\label{sec:reuse-evidence}

Once the testing level $\alpha_t$ has been determined, the next step is to construct the test statistic $M_{t,m}^{(\beta)}$. We first correct for bias from reusing validation samples. Based on the sampling procedure in Section~\ref{sec:setup}, we define the following index sets to characterize overlap across validation rounds. Recall that $K_{t,i}$ denotes the index of the $i$th validation sample in round $t$. Let $\mathcal R_{t,i-1}$ denote the set of indices not sampled in the first $i-1$ draws of round $t$, and let $\mathcal I_{t,i-1}$ denote the cumulative set of indices used in validation across the first $t-1$ rounds and the first $i-1$ draws of round $t$. For each draw, define
\begin{equation}
Q_{t,i}
=
\ind\{K_{t,i}\in\mathcal I_{t,i-1}\},
\qquad
\zeta_{t,i}
=
\Prb(Q_{t,i}=1\mid\mathcal H_{t,i-1})
=
\frac{|\mathcal I_{t,i-1}
\cap\mathcal R_{t,i-1}|}
{|\mathcal R_{t,i-1}|}.
\label{eq:exposure-rate}
\end{equation}
Here, $Q_{t,i}$ indicates whether the current sample has previously been used in validation, while $\zeta_{t,i}$ represents its
probability under uniform sampling.

\begin{algorithm}[!htbp]
\caption{\TRUST{}}
\label{alg:TRUST}
\raggedright
\renewcommand{\algorithmicrequire}{\textbf{Input:}}
\algsetup{indent=1.2em}
\begin{algorithmic}[1]
\REQUIRE Number of rounds $T$, validation pool $\mathcal V_N$, sample size $1\le n\le N$, initial incumbent $\Theta_0$, proposer $Q$, allocator $\Psi$, budget $q$, margin $\varepsilon$, strength $\beta$, $\alpha_{\max}$, $0<\rho_{\min}<\rho_{\max}<1$.
\STATE Initialize $B_1\gets q$, $R_0\gets0$, and $\mathcal I_{1,0}\gets\varnothing$.
\FOR{$t=1,\ldots,T$}
    \STATE Initialize $\tau_t \gets n$, $A_t\gets0$, $\mathcal R_{t,0}\gets\{1,\ldots,N\}$, and $M^{(\beta)}_{t,0}\gets1$.
    \STATE Propose $\Thetac_t\sim Q(\cdot\mid\F_{t-1})$ and construct context $C_t$.
    \STATE Set $w_t\gets\Psi(C_t)$ and $\rho_t\gets\operatorname{clip}\bigl(w_t/(T-t+1),\rho_{\min},\rho_{\max}\bigr)$.
    \STATE Allocate $\alpha_t\gets\min\{\alpha_{\max},\rho_t B_t(R_{t-1}+1)\}$.
    \STATE Update $B_{t+1}\gets B_t-\alpha_t/(R_{t-1}+1)$.
    \FOR{$i=1,\ldots,n$}
        \STATE Draw $K_{t,i}$ uniformly from $\mathcal R_{t,i-1}$ and evaluate $D_{t,i}$ on $X_{t,i}=Z_{K_{t,i}}$ by~\eqref{eq:paired-difference}.
        \STATE Compute $M^{(\beta)}_{t,i}$ using $D_{t,i}$, $\mathcal R_{t,i-1}$ and $\mathcal I_{t,i-1}$ by~\eqref{eq:exposure-rate} and~\eqref{eq:reuse-bet-y}.
        \STATE Update $\mathcal R_{t,i}\gets\mathcal R_{t,i-1}\setminus\{K_{t,i}\}$ and $\mathcal I_{t,i}\gets\mathcal I_{t,i-1}\cup\{K_{t,i}\}$.
        \STATE \textbf{if} $M^{(\beta)}_{t,i}\ge1/\alpha_t$ \textbf{then} $A_t\gets1$, $\tau_t\gets i$, break.
    \ENDFOR
    \STATE Update $\mathcal I_{t+1,0}\gets\mathcal I_{t,\tau_t}$.
    \STATE Set $\Theta_t\gets\Thetac_t$ if $A_t=1$, and $\Theta_t\gets\Theta_{t-1}$ otherwise.
    \STATE Update $R_t\gets R_{t-1}+A_t$.
\ENDFOR
\RETURN $\Theta_T$.
\end{algorithmic}
\end{algorithm}

We rescale the score difference in \eqref{eq:paired-difference}
from $[-1,1]$ to $[0,1]$ by setting $Y_{t,i}=(1+D_{t,i})/2$, which allows us to use standard e-value constructions for bounded observations \citep{waudbysmith2024estimating,ramdas2023game}, and the corresponding null boundary
in \eqref{eq:noninferiority-test} becomes
$p_\varepsilon=(1-\varepsilon)/2$. We choose $\beta\in[0,1-p_\varepsilon]$
as the correction strength for bias from data reuse and weight this strength by the exposure probability $\zeta_{t,i}$ to obtain the corrected null boundary
\(
u_{t,i}^{(\beta)}
=
p_\varepsilon+\zeta_{t,i}\beta.
\) The sequential e-value evidence is accumulated by setting $M_{t,0}^{(\beta)}=1$ and
\begin{equation}
M_{t,m}^{(\beta)}
=
\prod_{i=1}^{m}
\left[
1+\eta_{t,i}
\frac{Y_{t,i}-u_{t,i}^{(\beta)}}{u_{t,i}^{(\beta)}}
\right],
\qquad m=1,\ldots,n.
\label{eq:reuse-bet-y}
\end{equation}
The betting fraction $\eta_{t,i}\in[0,1)$ is chosen based on the information in $\mathcal H_{t,i-1}$ before observing $Y_{t,i}$ and controls how strongly the current observation changes the evidence. It can be selected using standard adaptive betting strategies \citep{waudbysmith2024estimating,orabona2024tight,shekhar2024nonparametric}. The complete procedure is presented in Algorithm~\ref{alg:TRUST}.

\section{Theoretical Analysis}
\label{sec:theory}
We now establish statistical guarantees of our framework, including the validity of the e-value, FER control, power consistency to accept improvements, and a lower bound on the final gain.

\begin{assumption}
\label{ass:exposure-bias}
For any $t=1,\ldots,T$ and $i=1,\ldots,n$, suppose 
\begin{equation}
\E[Y_{t,i}\mid\mathcal H_{t,i-1},Q_{t,i}=1]
\le p_\varepsilon+\beta,
\quad\text{if }H_t^0\text{ is true and }\zeta_{t,i}>0.
\label{eq:exposed-mean-bound}
\end{equation}
\end{assumption}
Reusing validation feedback to construct or select later candidates can lead to optimistic bias in their estimated performance, which is widely studied in adaptive data analysis \citep{dwork2015reusable,rogers2016maxinformation}. We use $\beta$ to represent the strength of the correction for data reuse. Assumption~\ref{ass:exposure-bias} connects this correction to our testing by bounding the conditional mean in the e-value construction. We discuss the relationship between $\beta$ and our algorithm in detail in Appendix~\ref{app:suppforoverlap}.

TRUST may accept the candidate early when the testing threshold is reached and the number of observations is a random variable determined by the observed evidence rather than a fixed number. We formalize this as a stopping time \citep{durrett2019probability}. A random variable $\tau_t\in\{1,\ldots,n\}$ is a stopping time with respect to the filtration $(\mathcal H_{t,i})_{i=0}^{n}$ if $\{\tau_t\le i\}\in\mathcal H_{t,i}$ for every $i=0,\ldots,n$. We then establish the following conditional validity result.
\begin{theorem}
\label{thm:evidence}
Fix any round $t\in\{1,\ldots,T\}$. Under Assumption~\ref{ass:exposure-bias}, for any $m=1,\ldots,n$, the statistic $M^{(\beta)}_{t,m}$ defined in \eqref{eq:reuse-bet-y} satisfies
\begin{equation}
\E[M^{(\beta)}_{t,m}\mid\Fpre_t]\le1,
\quad\text{if }H_t^0\text{ is true}.
\label{eq:conditional-evalidity}
\end{equation}
This inequality also holds with $m$ replaced by any stopping time $\tau_t\le n$ with respect to $(\mathcal H_{t,m})_{m=0}^{n}$.
\end{theorem}
Theorem~\ref{thm:evidence} shows that our test statistic satisfies the conditional validity requirement for e-values described in the statistical literature \citep{vovk2021evalues,ramdas2023game}. More importantly, validity at arbitrary stopping times $\tau_t\le n$ justifies the acceptance rule in Algorithm~\ref{alg:TRUST}, which stops at $\tau_t=\min\bigl\{n,\min\{i\in\{1,\ldots,n\}:M^{(\beta)}_{t,i}\ge1/\alpha_t\}\bigr\}$. The conditional validity also extends to mixtures of evidence processes using different betting fractions. See Corollary~\ref{cor:mixture} in the appendix. This validity result provides the basis for the following theorem on FER control in RSI.

\begin{theorem}
\label{thm:fer}
Fix a horizon $T\ge1$ and an FER budget $q\in(0,1)$. Under Assumption~\ref{ass:exposure-bias}, the false evolution rate of Algorithm~\ref{alg:TRUST} satisfies
\begin{equation}
\label{th:fer}
\FER_\varepsilon(T)\le q.
\end{equation}
\end{theorem}

Theorem~\ref{thm:fer} controls the expected proportion of unacceptable
updates among all accepted updates over the entire self-improvement process with $T$ rounds. The same bound in \eqref{th:fer} holds with $T$ replaced by any stopping time $\tau\le T$ with respect to the self-improvement history $(\F_t)_{t=0}^{T}$, as established in Appendix~\ref{app:outer-stopping}. Thus, the process may be stopped based on the observed self-improvement
history without losing FER control. In addition, the FER guarantee does not rely on the LLM correctly identifying beneficial candidates. The LLM uses context to allocate more FER budget to promising candidates and less to unpromising ones, safely improving the detection of beneficial updates.

Let $I_t=\ind\{\Delta_t>0\}$, $U_T=\sum_{t\le T}I_t$, and $S_T=\sum_{t\le T}I_tA_t$. The true evolution rate (TER) is
\begin{equation}
\TER(T)=\E\left[\frac{S_T}{U_T\vee1}\right],
\label{eq:ter-definition}
\end{equation}
the expected proportion of improving candidates that are accepted \citep{gadbury2004power}.

\begin{assumption}
\label{ass:resource-consistency}
Fix $T$ and $\varepsilon\in(0,1)$, and let $N,n\to\infty$.
Suppose that the following conditions are satisfied:
(a) There exists a deterministic sequence $r_{N,n}\to0$ such that
$\Prb(U_T=0)\le r_{N,n}$.
(b) $n/N\to0$.
(c) There exist constants $\eta_-$ and $\eta_+$ such that, for $t=1,\ldots,T$ and $i=1,\ldots,n$, the betting fractions satisfy $0<\eta_-\le\eta_{t,i}\le\eta_+<2\varepsilon/(1+\varepsilon)$.
\end{assumption}
Assumption~\ref{ass:resource-consistency}(a) requires the probability that the proposer fails to generate any improvement to vanish as the sample size grows. As \(n\to\infty\), the proposer also receives more historical feedback, making it more likely to produce at least one effective improvement. Assumption~\ref{ass:resource-consistency}(b) controls overlap by requiring \(n/N\to0\), so the bias caused by overlap becomes very small. Assumption~\ref{ass:resource-consistency}(c) can be satisfied in practice by choosing the betting fraction $\eta_{t,i}$ within the specified range.

\begin{theorem}
\label{thm:ter}
Under Assumption~\ref{ass:resource-consistency}, there exists a constant
$\kappa>0$ such that for sufficiently large
$N$ and $n$,
\begin{equation}
\label{eq:ter-consistency}
\operatorname{TER}(T)
\ge
1-r_{N,n}-Te^{-\kappa n}.
\end{equation}
Hence, $\operatorname{TER}(T)\to1$ as $N,n\to\infty$.
\end{theorem}
Theorem~\ref{thm:ter} provides a lower bound on TER through two terms:
$r_{N,n}$ bounds the probability that no improving candidate is proposed,
while $Te^{-\kappa n}$ bounds the contribution of gate misses. Both terms vanish as the sample size grows with $T$ fixed. It also reveals a theoretical role of noninferiority: when \(\varepsilon=0\), the upper bound \(2\varepsilon/(1+\varepsilon)\) in Assumption~\ref{ass:resource-consistency} is zero, then no positive betting fraction satisfies the condition. Under this argument, a positive \(\varepsilon\) is needed to obtain \(\operatorname{TER}(T)\to1\).

We next evaluate the final incumbent through its gain $G(T)$ defined in \eqref{eq:gain}. Let $\mathcal P_T = \sum_{t=1}^{T}\E[(\Delta_t)_+]$ denote the expected total positive effect of the proposed candidates. 

\begin{theorem}
\label{thm:utility}
Under Assumptions~\ref{ass:exposure-bias}
and~\ref{ass:resource-consistency}, there exists a constant
$\kappa>0$ such that for sufficiently
large $N$ and $n$,
\begin{equation}
\begin{aligned}
G(T)\ge{}&
\bigl(1-e^{-\kappa n}\bigr)\mathcal P_T-
\sum_{t=1}^{T}
\E\!\left[
A_t\min\{(-\Delta_t)_+,\varepsilon\}
+
\alpha_t(-\Delta_t-\varepsilon)_+
\right].
\end{aligned}
\label{eq:deployment-gain-bound}
\end{equation}
\end{theorem}
The first term shows that the loss of positive improvement caused by statistical rejection vanishes exponentially with the validation sample size. The second term quantifies the downside from the noninferiority margin. The third term controls loss beyond the tolerance through the allocated testing levels \(\alpha_t\) and may be reduced when the LLM uses historical context to assign smaller testing levels to candidates with greater downside risk. Thus, TRUST asymptotically preserves beneficial updates while keeping harmful acceptance risk under statistical control. Additionally, to measure losses along the trajectory, we define population maximum drawdown (MDD) as $\operatorname{MDD}(T)=\max_{0\le r<s\le T}\bigl(J(\Theta_r)-J(\Theta_s)\bigr)_+$. Appendix~\ref{app:gain-proof} derives an upper bound on expected MDD.

\section{Numerical Experiments}
\label{sec:harness}
\label{sec:experiments}
We compare the performance gains and acceptance risks of different gates in an editable self-improvement harness, and conduct ablation studies to examine the effects of TRUST's key parameters. Supplementary diagnostics and case studies in Appendices~\ref{app:hotpot-overfitting} and~\ref{app:apps-gate-case} show that higher validation scores do not necessarily imply better test performance.

Following modular agent design \citep{shang2025agentsquare,hu2025adas}, our editable harness $\mathsf H_t=(\mathsf R_t,\mathsf P_t,\mathsf M_t,\mathsf W_t)$ exposes 20 operators across four components: reasoning \citep{zhou2023least,fernando2024promptbreeder}, routing \citep{lewis2020rag,liu2024lost}, memory \citep{shinn2023reflexion,zhao2024expel}, and workflow \citep{zhang2025aflow}. Appendix~\ref{app:scheme820} details operator designs.

\subsection{Main gate comparison}
\label{sec:main-gate-comparison}

\paragraph{Datasets, models and metrics.}
We evaluate the benchmarks HotpotQA \citep{yang2018hotpotqa}, DROP \citep{dua2019drop}, and APPS \citep{hendrycks2021apps} with the models Qwen-3.8-Flash and DeepSeek-V4-Flash. For each dataset, the validation sets contain a total of 200 examples, with 300 held-out examples used only as the test set. We use additional 20 examples to construct the initial information $\mathcal F_0$. Test scores are measured by Joint F1 for HotpotQA, official token F1 for DROP, and the test case pass rate for APPS. We report gain and MDD in percentage points (pp). For a stricter empirical diagnostic, we report \(\mathrm{FER}_{0}\), which counts all accepted non-improving updates.

\paragraph{Baselines.}
We compare \TRUST{} with four baselines over 30 rounds and 5 seeds. Greedy accepts candidates with higher scores on the  validation samples. PACE uses an e-process test to control false acceptance within a single round \citep{pace2026}. SGM-B and SGM-H use Bernstein and Hoeffding tests, respectively, with FWER control across rounds \citep{wu2025sgm}. For \TRUST{}, we set $q=0.30$, $\varepsilon=0.10$, and $\beta=0.01$. More implementation details appear in Appendix~\ref{app:main-experiments}.

\begingroup
\begin{table}[H]
\centering
\caption{Main comparison: DeepSeek-V4-Flash (left) and Qwen-3.8-Flash (right). Values are mean $\pm$ standard error. Methods with empirical $\mathrm{FER}_0>30\%$ are shown in gray to indicate high risk. Bold marks the best Gain, TER, and MDD results among the remaining methods within each setting.}
\label{tab:main96}
\scriptsize
\renewcommand{\arraystretch}{0.95}
\setlength{\tabcolsep}{1pt}
\newcommand{\mainresult}[2]{\mbox{#1\kern.25pt{\tiny$\pm$#2}}}
\begin{tabular*}{\linewidth}{@{\extracolsep{\fill}}llrrrr@{\hspace{7pt}}rrrr@{}}
\toprule
& & \multicolumn{4}{c}{\bfseries DeepSeek-V4-Flash} & \multicolumn{4}{c}{\bfseries Qwen-3.8-Flash} \\
\cmidrule(lr){3-6}\cmidrule(l){7-10}
Dataset & Gate & Gain (pp) $\uparrow$ & FER$_0$ (\%) $\downarrow$ & TER (\%) $\uparrow$ & MDD (pp) $\downarrow$ & Gain (pp) $\uparrow$ & FER$_0$ (\%) $\downarrow$ & TER (\%) $\uparrow$ & MDD (pp) $\downarrow$ \\
\midrule

\multirow{5}{*}{HotpotQA}
& \textcolor{black!50}{Greedy}
& \color{black!50} \mainresult{7.87}{0.55}
& \color{black!50} \mainresult{64.67}{3.28}
& \color{black!50} \mainresult{40.03}{5.06}
& \color{black!50} \mainresult{4.96}{0.93}
& \color{black!50} \mainresult{20.66}{2.45}
& \color{black!50} \mainresult{44.67}{5.23}
& \color{black!50} \mainresult{58.40}{9.39}
& \color{black!50} \mainresult{4.01}{0.49} \\

& PACE
& \mainresult{10.15}{0.75}
& \mainresult{0.00}{0.00}
& \mainresult{12.63}{3.65}
& \mainresult{\textbf{0.00}}{0.00}
& \mainresult{26.42}{1.60}
& \mainresult{6.67}{6.67}
& \mainresult{16.31}{2.94}
& \mainresult{0.06}{0.06} \\

& SGM-B
& \mainresult{0.00}{0.00}
& \mainresult{0.00}{0.00}
& \mainresult{0.00}{0.00}
& \mainresult{\textbf{0.00}}{0.00}
& \mainresult{0.00}{0.00}
& \mainresult{0.00}{0.00}
& \mainresult{0.00}{0.00}
& \mainresult{\textbf{0.00}}{0.00} \\

& SGM-H
& \mainresult{0.00}{0.00}
& \mainresult{0.00}{0.00}
& \mainresult{0.00}{0.00}
& \mainresult{\textbf{0.00}}{0.00}
& \mainresult{5.77}{5.77}
& \mainresult{0.00}{0.00}
& \mainresult{1.25}{1.25}
& \mainresult{\textbf{0.00}}{0.00} \\

& TRUST
& \mainresult{\textbf{11.58}}{1.30}
& \mainresult{10.00}{10.00}
& \mainresult{\textbf{18.11}}{8.39}
& \mainresult{0.02}{0.02}
& \mainresult{\textbf{27.72}}{0.81}
& \mainresult{0.00}{0.00}
& \mainresult{\textbf{20.82}}{3.69}
& \mainresult{\textbf{0.00}}{0.00} \\

\addlinespace[3pt]

\multirow{5}{*}{DROP}
& \textcolor{black!50}{Greedy}
& \color{black!50} \mainresult{36.24}{2.19}
& \color{black!50} \mainresult{35.28}{1.68}
& \color{black!50} \mainresult{60.12}{7.77}
& \color{black!50} \mainresult{5.38}{1.58}
& \color{black!50} \mainresult{15.15}{2.22}
& \color{black!50} \mainresult{37.38}{6.04}
& \color{black!50} \mainresult{56.33}{5.65}
& \color{black!50} \mainresult{3.37}{1.03} \\

& PACE
& \mainresult{29.70}{0.59}
& \mainresult{0.00}{0.00}
& \mainresult{6.68}{1.67}
& \mainresult{\textbf{0.00}}{0.00}
& \mainresult{13.22}{3.77}
& \mainresult{0.00}{0.00}
& \mainresult{5.23}{1.78}
& \mainresult{\textbf{0.00}}{0.00} \\

& SGM-B
& \mainresult{0.00}{0.00}
& \mainresult{0.00}{0.00}
& \mainresult{0.00}{0.00}
& \mainresult{\textbf{0.00}}{0.00}
& \mainresult{0.00}{0.00}
& \mainresult{0.00}{0.00}
& \mainresult{0.00}{0.00}
& \mainresult{\textbf{0.00}}{0.00} \\

& SGM-H
& \mainresult{0.00}{0.00}
& \mainresult{0.00}{0.00}
& \mainresult{0.00}{0.00}
& \mainresult{\textbf{0.00}}{0.00}
& \mainresult{0.00}{0.00}
& \mainresult{0.00}{0.00}
& \mainresult{0.00}{0.00}
& \mainresult{\textbf{0.00}}{0.00} \\

& TRUST
& \mainresult{\textbf{31.47}}{1.42}
& \mainresult{0.00}{0.00}
& \mainresult{\textbf{11.93}}{1.29}
& \mainresult{\textbf{0.00}}{0.00}
& \mainresult{\textbf{17.22}}{1.49}
& \mainresult{0.00}{0.00}
& \mainresult{\textbf{10.76}}{2.82}
& \mainresult{\textbf{0.00}}{0.00} \\

\addlinespace[3pt]

\multirow{5}{*}{APPS}
& \textcolor{black!50}{Greedy}
& \color{black!50} \mainresult{33.89}{2.71}
& \color{black!50} \mainresult{41.08}{4.31}
& \color{black!50} \mainresult{50.83}{2.89}
& \color{black!50} \mainresult{8.48}{2.23}
& \color{black!50} \mainresult{28.03}{0.32}
& \color{black!50} \mainresult{50.44}{4.85}
& \color{black!50} \mainresult{38.02}{4.36}
& \color{black!50} \mainresult{3.59}{0.68} \\

& PACE
& \mainresult{28.68}{2.66}
& \mainresult{0.00}{0.00}
& \mainresult{5.19}{1.00}
& \mainresult{\textbf{0.00}}{0.00}
& \mainresult{26.24}{0.46}
& \mainresult{0.00}{0.00}
& \mainresult{4.44}{0.48}
& \mainresult{\textbf{0.00}}{0.00} \\

& SGM-B
& \mainresult{0.00}{0.00}
& \mainresult{0.00}{0.00}
& \mainresult{0.00}{0.00}
& \mainresult{\textbf{0.00}}{0.00}
& \mainresult{0.00}{0.00}
& \mainresult{0.00}{0.00}
& \mainresult{0.00}{0.00}
& \mainresult{\textbf{0.00}}{0.00} \\

& SGM-H
& \mainresult{0.00}{0.00}
& \mainresult{0.00}{0.00}
& \mainresult{0.00}{0.00}
& \mainresult{\textbf{0.00}}{0.00}
& \mainresult{10.84}{6.65}
& \mainresult{0.00}{0.00}
& \mainresult{2.62}{1.70}
& \mainresult{\textbf{0.00}}{0.00} \\

& TRUST
& \mainresult{\textbf{34.55}}{1.04}
& \mainresult{0.00}{0.00}
& \mainresult{\textbf{11.64}}{3.53}
& \mainresult{\textbf{0.00}}{0.00}
& \mainresult{\textbf{26.53}}{0.11}
& \mainresult{6.67}{6.67}
& \mainresult{\textbf{12.52}}{3.42}
& \mainresult{0.27}{0.27} \\

\bottomrule
\end{tabular*}
\vspace{0pt}
\end{table}
\endgroup

\TRUST{} achieves high scores in all six settings and the highest scores in four of them (Table~\ref{tab:main96}). It maintains FER$_0$ below 30\% in all six settings, whereas Greedy exceeds 30\% in every setting. The highest MDD for \TRUST{} is only 0.27~pp, whereas Greedy reaches 8.48~pp. Appendix~\ref{app:apps-gate-case} presents a case study of Greedy that improves validation performance but harms test performance.

Among the remaining methods with $\mathrm{FER}_0\leq30\%$, SGM-B never accepts an update and SGM-H accepts only three updates across all runs, so they show no score improvement in most settings. These results suggest that both methods for FWER control across multiple rounds are conservative in practice. Relative to PACE, \TRUST{} achieves higher scores and TER in all six settings, with FER$_0$ lower in one setting, equal in three, and higher in two. 

\begin{figure}[!http]
\centering
\includegraphics[width=\linewidth]{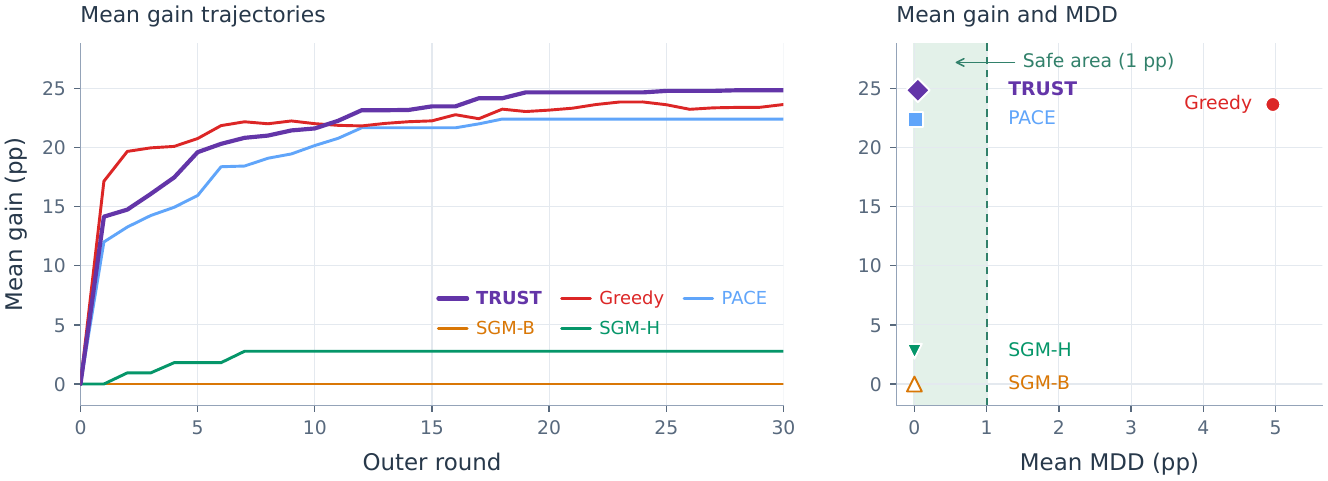}
\caption{Left: mean gain trajectories. Right: mean final gain versus mean MDD, with shading marking the 1~pp empirical MDD reference region. Results are averaged across six settings.}
\label{fig:gate-comparison-gain-and-drawdown}
\end{figure}

The left panel of Figure~\ref{fig:gate-comparison-gain-and-drawdown} shows the mean gain trajectories over rounds. Greedy achieves rapid initial gains but exhibits subsequent declines. \TRUST{} overtakes Greedy in later rounds and preserves its accumulated gains, achieving the highest final mean gain. The right panel compares mean final gain and mean MDD. We mark the region with mean MDD below 1~pp as an empirical safe area. \TRUST{} lies within this region while achieving the highest among all compared methods, striking a balance between performance gain and stability in RSI.

\FloatBarrier
\subsection{Ablation study}
\label{sec:sensitivity}

We next conduct ablation studies on HotpotQA/DeepSeek-V4-Flash. We vary three key parameters: the FER budget $q$, noninferiority margin $\varepsilon$, and overlap correction strength $\beta$. We also compare TRUST with and without LLM guidance. Appendix~\ref{app:sensitivity} provides the detailed settings and results.

\begin{figure}[!htbp]
\includegraphics[width=\linewidth]{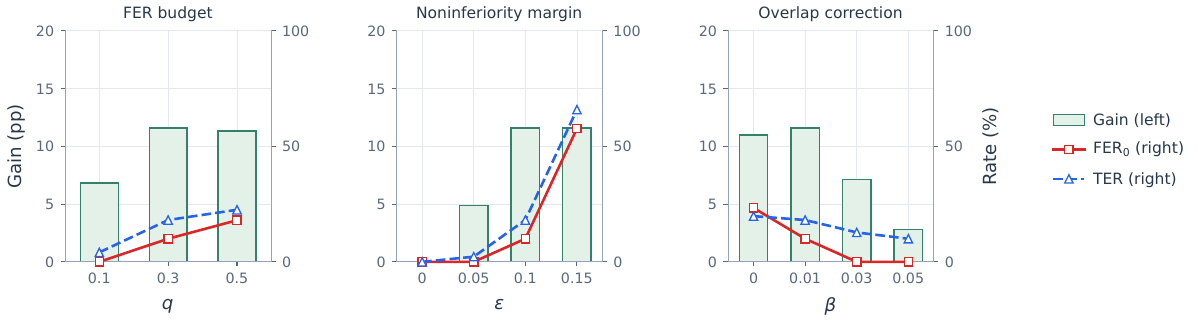}
\caption{Bars show mean gain, and lines show FER$_0$ and TER. We vary one parameter at a time, keeping the others at the default values $(q,\varepsilon,\beta)=(0.30,0.10,0.01)$.}
\label{fig:parameter-sensitivity}
\end{figure}

\begin{wrapfigure}{r}{0.4\textwidth}
\centering
\includegraphics[width=\linewidth]{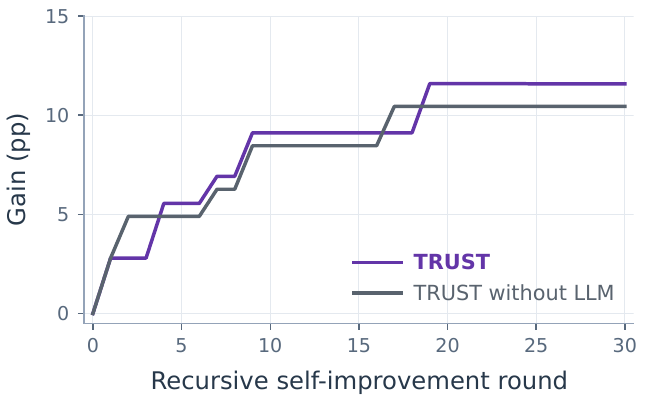}
\caption{Mean gain trajectories with and without LLM allocation.}
\label{fig:allocator-ablation}
\end{wrapfigure}

Figure~\ref{fig:parameter-sensitivity} shows that larger $q$ and $\varepsilon$ make \TRUST{} more aggressive, with higher TER and FER$_0$. Larger $\beta$ makes it more conservative: FER$_0$ drops from 23.33\% at $\beta=0$ to 0\% at $\beta=0.03$. These trends are consistent with the definitions of the three parameters. In each ablation study, the gain first increases, then decreases or holds steady. For $\beta$, it peaks at 11.58~pp when $\beta=0.01$, compared with 10.97~pp without correction, suggesting that an appropriate overlap correction strength can be beneficial in practice. Appendix~\ref{app:parasensitivity} provides additional trajectory results and a more detailed analysis. These results highlight the importance of balancing conservative and aggressive acceptance to improve performance.

We replace the LLM allocator with the default multiplier $w_t \equiv 1$ while keeping other parameters in TRUST unchanged. Figure~\ref{fig:allocator-ablation} shows that TRUST achieves a higher final gain, corresponding to final scores of 44.91 $\pm$ 1.36\% for TRUST and 43.78 $\pm$ 1.18\% for TRUST without LLM guidance. This suggests a performance benefit from using context to guide the allocation of testing levels. We provide a detailed case study in Appendix~\ref{sec:allocator-case-study} showing how the LLM analyzes context and assigns different LLM multipliers to influence acceptance decisions.

\FloatBarrier

\section{Conclusion and Limitations}
\TRUST{} applies online multiple hypothesis testing to control FER across rounds in self-improving agents, with LLM guidance and interpretable parameters for adjusting gate behavior. Our theoretical analysis and experiments show that the framework helps balance useful improvement and acceptance error control under limited validation resources. It extends existing statistical methods and provides a flexible statistical foundation for future work on recursive self-improvement. We discuss this point further in Appendix~\ref{app:suppforStatistical}. Overall, we show that \TRUST{} strikes a balance between statistical reliability and performance gains in recursive self-improvement.

The present study does not address how to identify an exact value of overlap correction strength from data or how to optimally tune other parameters. The noninferiority margin provides an interpretable tolerance for loss, but the theoretical guarantee can be conservative and only partly explains practical behavior. The LLM allocator can occasionally make mistakes when judging candidate updates, which may lead to poor choices of LLM multipliers. Optimizing the gate to maximize gain while keeping safety metrics within target ranges is also a promising direction. Future work can address these limitations to make the recursive self-improvement process more trustworthy and effective.

\clearpage
\nocite{shafer2021testing,shafer2011test}
\bibliographystyle{iclr2027_conference}
\bibliography{reference}

\clearpage
\appendix
\appendix
\section*{Appendix}

\section{Supplementary Discussion}
\label{app:supplementary-discussion}

\subsection{Interpretation of the overlap correction}
\label{app:suppforoverlap}
We do not develop a complete model of validation data reuse in this work. Detailed studies of adaptive data analysis can be found in prior work \citep{dwork2015reusable,rogers2016maxinformation,
russo2016controlling,xu2017information}. Instead, we state sufficient conditions for valid e-values and focus on a framework for FER control across RSI rounds. The interpretation of $\beta$ is related to how historical validation information, including feedback summarized in $C_t$, is used in candidate construction and selection. Greater reliance on information from previously evaluated samples may increase optimism when those samples are reused. For example, later rounds may use a few failed cases from the previous round or summaries of common error types. These forms of feedback reveal different information about the validation data and may lead to different degrees of optimism bias in subsequent evaluation. The parameter $\beta$ specifies an allowance for this optimism under the null hypothesis. A stronger correction requires stronger validation results to reach the acceptance threshold, making the gate more conservative and potentially limiting both false acceptances and the acceptance of useful updates. Our sensitivity results in Section~\ref{sec:sensitivity} and Appendix~\ref{app:sensitivity} support the practical value of an appropriate correction strength. In the reported ablation, an appropriate $\beta$ achieves a higher mean final gain than no correction, whereas larger values make the gate overly conservative and reduce final gains.

\subsection{Relation to Existing Statistical Methods}
\label{app:suppforStatistical}
TRUST adapts FDR control in multiple hypothesis testing to the acceptance of candidates in self-evolving agents. It provides a framework covering the validation protocol, correction for data reuse, noninferiority testing, construction of test statistics, and allocation of testing levels. Within this framework, an LLM analyzes the candidate and historical context to guide allocation, while statistical guarantees hold under the stated assumptions. Existing statistical studies \citep{zhang2025egai,chen2020contextual} compare testing procedures on specified hypotheses and observation streams in their numerical experiments. In our RSI setting, the proposer constructs candidates using historical information. Different gates can produce different acceptance decisions and incumbents, leading to different histories and subsequent candidates. Our experiments follow this closed loop and use update operators motivated by prior work on RSI agents. This allows us to compare how different gates affect both acceptance decisions and subsequent search on real tasks. Our theoretical analysis follows the same adaptive process and connects a lower bound on final gain to the testing levels and other parameters of the gate. It thus provides a statistical foundation for analyzing both acceptance risk and improvement along the trajectory induced
by the gate.

\section{Proof of Conditional Evidence Validity}

\subsection{Proof of Theorem~\ref{thm:evidence}}
\label{app:evidence-proof}

In each round, the sampling plan is to draw $n$ tasks from the fixed validation pool of size $N$, where $n\le N$. All $N$ sample indices are selectable at the start of the round, so $\mathcal R_{t,0}=\{1,2,\ldots,N\}$. At draw $i$, we select $K_{t,i}$ uniformly from the remaining set $\mathcal R_{t,i-1}$ and remove it, giving $\mathcal R_{t,i}=\mathcal R_{t,i-1}\setminus\{K_{t,i}\}$. Thus, a sample can be drawn at most once within a round but may be used again in later rounds.

Let $L_{t,i}\in\{0,1\}$ be the $\mathcal H_{t,i-1}$-measurable indicator that draw $i$ is made.
\[
L_{t,i}=
\begin{cases}
1, & \text{if draw $i$ is made in round $t$},\\
0, & \text{otherwise}.
\end{cases}
\]
Within each round, sampling proceeds consecutively and does not continue after stopping. If $\alpha_t=0$, no validation samples are drawn.

Each round uses at most $n$ samples. The bounds on the betting fractions in Assumption~\ref{ass:resource-consistency}(c) apply on draws that are made ($L_{t,i}=1$). If sampling stops early, we set the variables for the remaining indices to zero as a notational convention for the proofs:
\[
Y_{t,i}=Q_{t,i}=\zeta_{t,i}=\eta_{t,i}=0
\qquad\text{when }L_{t,i}=0.
\]
After stopping, no more samples are used, and the history and record of previously used samples stay unchanged. 
When a sample is drawn, $Q_{t,i}$ equals one with conditional probability $\zeta_{t,i}$. Hence,
\[
\E[Q_{t,i}\mid\mathcal H_{t,i-1}]=\zeta_{t,i}.
\]
After stopping, both $Q_{t,i}$ and $\zeta_{t,i}$ are set to zero, so this equality still holds.

\begin{lemma}
\label{lem:bounded-mean-bet}
Let $(\mathcal H_i)_{i=0}^{n}$ be a filtration, $Y_i\in[0,1]$ be $\mathcal H_i$-measurable, and $u_i\in(0,1]$ and $\eta_i\in[0,1)$ be $\mathcal H_{i-1}$-measurable for each $i=1,\ldots,n$. If
\[
\E[Y_i\mid\mathcal H_{i-1}]\le u_i,
\]
then
\[
M_0=1,
\qquad
M_m=\prod_{i=1}^m
\left[1+\eta_i\frac{Y_i-u_i}{u_i}\right]
\]
is a nonnegative supermartingale. Moreover, for every stopping time $\tau\le n$,
$\E[M_\tau\mid\mathcal H_0]\le1$.
\end{lemma}

\begin{proof}
Write $G_i=1+\eta_i(Y_i-u_i)/u_i$. Since $Y_i\ge0$,
$G_i\ge1-\eta_i>0$. Since $u_i$ and $\eta_i$ are $\mathcal H_{i-1}$-measurable, we have
\[
\E[G_i\mid\mathcal H_{i-1}]
=1+\frac{\eta_i}{u_i}
\left(\E[Y_i\mid\mathcal H_{i-1}]-u_i\right)
\le1.
\]
By the product definition, $M_i=M_{i-1}G_i$ and $M_{i-1}=\prod_{j=1}^{i-1}G_j$. For $j<i$, each $G_j$ is $\mathcal H_j$-measurable, with $\mathcal H_j\subseteq\mathcal H_{i-1}$. Thus, $M_{i-1}$ is $\mathcal H_{i-1}$-measurable and can be taken outside the conditional expectation:
\begin{align*}
\E[M_i\mid\mathcal H_{i-1}]
&=\E[M_{i-1}G_i\mid\mathcal H_{i-1}]\\
&=M_{i-1}\E[G_i\mid\mathcal H_{i-1}]\\
&=M_{i-1}\left[1+\frac{\eta_i}{u_i}
\left(\E[Y_i\mid\mathcal H_{i-1}]-u_i\right)\right]\\
&\le M_{i-1}.
\end{align*}
Taking expectations and iterating from $M_0=1$ gives $\E[M_i]\le1$, so each $M_i$ is integrable. Together with nonnegativity and $\mathcal H_i$-measurability, this proves the supermartingale claim.

For bounded $\tau$, use the finite pathwise decomposition
\[
M_\tau=M_0+
\sum_{i=1}^{n}\ind\{\tau\ge i\}(M_i-M_{i-1}).
\]
The event $\{\tau\ge i\}=\{\tau\le i-1\}^c$ belongs to $\mathcal H_{i-1}$. Hence every stopped increment has nonpositive conditional mean:
\begin{align*}
&\E\left[
\ind\{\tau\ge i\}(M_i-M_{i-1})
\middle|\mathcal H_{i-1}
\right]\\
&\qquad=
\ind\{\tau\ge i\}
\left(\E[M_i\mid\mathcal H_{i-1}]-M_{i-1}\right)
\le0.
\end{align*}
Since $\mathcal H_0\subseteq\mathcal H_{i-1}$, the tower property implies that each stopped increment also has nonpositive conditional expectation given $\mathcal H_0$. Taking conditional expectations given $\mathcal H_0$ in the decomposition of $M_\tau$ above, we obtain
\begin{align*}
\E[M_\tau\mid\mathcal H_0]
&=M_0+\sum_{i=1}^{n}
\E\left[\ind\{\tau\ge i\}(M_i-M_{i-1})\mid\mathcal H_0\right]\\
&\le M_0=1.
\end{align*}
\end{proof}

\begin{proof}[Proof of Theorem~\ref{thm:evidence}]
Fix outer round $t$. Recall that
\begin{align*}
Y_{t,i}&=\frac{1+D_{t,i}}{2},
&\Delta_t&=J(\Thetac_t)-J(\Theta_{t-1}),\\
Q_{t,i}&=\ind\{K_{t,i}\in\mathcal I_{t,i-1}\},
&\zeta_{t,i}&=\Prb(Q_{t,i}=1\mid\mathcal H_{t,i-1}).
\end{align*}

Both $\mathcal R_{t,i-1}$ and $\mathcal I_{t,i-1}$ are $\mathcal H_{t,i-1}$-measurable. Uniform sampling from the remaining set gives
\begin{align*}
\Prb(Q_{t,i}=1\mid\mathcal H_{t,i-1})
&=\sum_{k\in\mathcal R_{t,i-1}}
\ind\{k\in\mathcal I_{t,i-1}\}
\frac1{|\mathcal R_{t,i-1}|}\\
&=\frac{|\mathcal I_{t,i-1}\cap
\mathcal R_{t,i-1}|}{|\mathcal R_{t,i-1}|}
=\zeta_{t,i}.
\end{align*}

We bound the contributions of unexposed and reused samples separately. The history depends only on accessed sample information and randomization independent of unexposed sample values. Since the pool is i.i.d., unexposed samples retain distribution $\mathcal D$ conditional on $\mathcal H_{t,i-1}$. Thus, for each draw,
\begin{equation}
\E[Y_{t,i}(1-Q_{t,i})\mid\mathcal H_{t,i-1}]
=(1-\zeta_{t,i})\frac{1+\Delta_t}{2}.
\label{eq:unexposed-mean}
\end{equation}
Here, $1-Q_{t,i}$ retains only the unexposed contribution, and $1-\zeta_{t,i}$ is the probability of drawing an unexposed sample.

For reused samples, Assumption~\ref{ass:exposure-bias} allows an additional $\beta$ above the null boundary $p_\varepsilon=(1-\varepsilon)/2$. On null histories with $\zeta_{t,i}>0$, we have
\begin{align*}
\E[Y_{t,i}Q_{t,i}\mid\mathcal H_{t,i-1}]
&=\zeta_{t,i}\,\E[Y_{t,i}\mid\mathcal H_{t,i-1},Q_{t,i}=1]\\
&\le\zeta_{t,i}(p_\varepsilon+\beta).
\end{align*}
When $\zeta_{t,i}=0$, conditional monotonicity instead gives
\[
0\le\E[Y_{t,i}Q_{t,i}\mid\mathcal H_{t,i-1}]
\le\E[Q_{t,i}\mid\mathcal H_{t,i-1}]=0,
\]
so the same weighted bound holds. Combining this bound with Equation~\eqref{eq:unexposed-mean} and $(1+\Delta_t)/2\le p_\varepsilon$ on $H_t^0$ yields
\begin{align*}
\E[Y_{t,i}\mid\mathcal H_{t,i-1}]
&=\E[Y_{t,i}Q_{t,i}\mid\mathcal H_{t,i-1}]+\E[Y_{t,i}(1-Q_{t,i})\mid\mathcal H_{t,i-1}]\\
&\le\zeta_{t,i}(p_\varepsilon+\beta)
 +(1-\zeta_{t,i})\frac{1+\Delta_t}{2}\\
&\le p_\varepsilon+\zeta_{t,i}\beta.
\end{align*}
The correction $\zeta_{t,i}\beta$ therefore weights the overlap correction by the probability of drawing a reused sample. This also covers $\zeta_{t,i}=1$, when the unexposed contribution vanishes.

When draw $i$ is not made, $L_{t,i}=0$ and $Y_{t,i}=Q_{t,i}=0$ by the convention above. Since $L_{t,i}$ is $\mathcal H_{t,i-1}$-measurable, the indicator $\ind\{L_{t,i}=0\}$ is also $\mathcal H_{t,i-1}$-measurable and can be taken outside the conditional expectation. Hence,
\begin{align*}
\ind\{L_{t,i}=0\}\E[Y_{t,i}\mid\mathcal H_{t,i-1}]
&=\E[\ind\{L_{t,i}=0\}Y_{t,i}\mid\mathcal H_{t,i-1}]=0,\\
\ind\{L_{t,i}=0\}\E[Q_{t,i}\mid\mathcal H_{t,i-1}]
&=\E[\ind\{L_{t,i}=0\}Q_{t,i}\mid\mathcal H_{t,i-1}]=0.
\end{align*}
On $\{L_{t,i}=0\}$, the indicator equals one, so $\E[Y_{t,i}\mid\mathcal H_{t,i-1}]=0$ and $\zeta_{t,i}=\E[Q_{t,i}\mid\mathcal H_{t,i-1}]=0$.

For draws that are made, the bounds for the unexposed and reused contributions together give $\E[Y_{t,i}\mid\mathcal H_{t,i-1}]\le u_{t,i}^{(\beta)}$ on $H_t^0$. For draws that are not made, either after early stopping or because $\alpha_t=0$, the zero conditional mean above is also at most $u_{t,i}^{(\beta)}=p_\varepsilon>0$. Recall that $N_t^{\varepsilon}=\ind\{\Delta_t\le-\varepsilon\}$. On $H_t^0$, multiplying by $N_t^{\varepsilon}=1$ preserves these bounds; outside $H_t^0$, multiplying by $N_t^{\varepsilon}=0$ makes both sides zero. Therefore, on all histories,
\[
N_t^{\varepsilon}\E[Y_{t,i}\mid\mathcal H_{t,i-1}]
\le N_t^{\varepsilon}u_{t,i}^{(\beta)}.
\]
The corrected boundary $u_{t,i}^{(\beta)}$ is $\mathcal H_{t,i-1}$-measurable and lies in $[p_\varepsilon,1]$, with $p_\varepsilon>0$. Write $G_{t,i}=1-\eta_{t,i}+\eta_{t,i}Y_{t,i}/u_{t,i}^{(\beta)}$. When draw $i$ is not made, we set $\eta_{t,i}=0$, so $G_{t,i}=1$. For every $i$, the bounds $0\le Y_{t,i}\le1$, $0\le\eta_{t,i}<1$, and $u_{t,i}^{(\beta)}\ge p_\varepsilon>0$ give
\begin{align*}
G_{t,i}&\ge1-\eta_{t,i}>0,\\
G_{t,i}&\le1-\eta_{t,i}+\frac{\eta_{t,i}}{p_\varepsilon}
=1+\eta_{t,i}(p_\varepsilon^{-1}-1)\\
&\le p_\varepsilon^{-1}.
\end{align*}
The last inequality uses $\eta_{t,i}\le1$ and $p_\varepsilon^{-1}-1\ge0$. Hence $0<G_{t,i}\le p_\varepsilon^{-1}$ almost surely.

Since each $G_{t,i}$ is $\mathcal H_{t,i}$-measurable, the product $M^{(\beta)}_{t,m}$ is $\mathcal H_{t,m}$-measurable. For $0\le m\le n\le N$, we have
\[
0\le N_t^{\varepsilon}M^{(\beta)}_{t,m}
\le M^{(\beta)}_{t,m}
=\prod_{i=1}^{m}G_{t,i}
\le p_\varepsilon^{-m}
\le p_\varepsilon^{-N}.
\]
Since $p_\varepsilon^{-N}$ is a finite deterministic constant, both $M^{(\beta)}_{t,m}$ and $N_t^{\varepsilon}M^{(\beta)}_{t,m}$ have finite expectations and are therefore integrable. After sampling stops, all remaining factors $G_{t,i}$ equal one, so both $M^{(\beta)}_{t,m}$ and $N_t^{\varepsilon}M^{(\beta)}_{t,m}$ remain unchanged.

Set $X_m=N_t^{\varepsilon}M^{(\beta)}_{t,m}$. Since $N_t^{\varepsilon}$ is $\mathcal H_{t,0}$-measurable, $X_m$ is $\mathcal H_{t,m}$-measurable; the bounds above also show that $X_m$ is nonnegative and integrable. To establish that $(X_m)_{m=0}^{n}$ is a supermartingale, it remains to prove
\[
\E[X_{m+1}\mid\mathcal H_{t,m}]\le X_m,
\qquad m=0,\ldots,n-1.
\]

Since $\mathcal H_{t,0}\subseteq\mathcal H_{t,i-1}$, $N_t^{\varepsilon}$ is also $\mathcal H_{t,i-1}$-measurable. The terms $1-\eta_{t,i}$ and $\eta_{t,i}/u_{t,i}^{(\beta)}$ are $\mathcal H_{t,i-1}$-measurable and bounded as well. We can therefore take these quantities outside the conditional expectation. Using the bound $N_t^{\varepsilon}\E[Y_{t,i}\mid\mathcal H_{t,i-1}]\le N_t^{\varepsilon}u_{t,i}^{(\beta)}$ established above, we obtain
\begin{align*}
\E[N_t^{\varepsilon}G_{t,i}\mid\mathcal H_{t,i-1}]
&=N_t^{\varepsilon}\E[G_{t,i}\mid\mathcal H_{t,i-1}]\\
&=N_t^{\varepsilon}(1-\eta_{t,i})
+\frac{\eta_{t,i}}{u_{t,i}^{(\beta)}}N_t^{\varepsilon}\E[Y_{t,i}\mid\mathcal H_{t,i-1}]\\
&\le N_t^{\varepsilon}(1-\eta_{t,i})
+\frac{\eta_{t,i}}{u_{t,i}^{(\beta)}}N_t^{\varepsilon}u_{t,i}^{(\beta)}\\
&=N_t^{\varepsilon}.
\end{align*}

The product before draw $i$ is $M^{(\beta)}_{t,i-1}=\prod_{j=1}^{i-1}G_{t,j}\ge0$. For $j<i$, each $G_{t,j}$ is $\mathcal H_{t,j}$-measurable, with $\mathcal H_{t,j}\subseteq\mathcal H_{t,i-1}$. Thus $M^{(\beta)}_{t,i-1}$ is $\mathcal H_{t,i-1}$-measurable and can also be taken outside the conditional expectation. Since $X_i=M^{(\beta)}_{t,i-1}N_t^{\varepsilon}G_{t,i}$, we have
\begin{align*}
\E[X_i\mid\mathcal H_{t,i-1}]
&=\E[M^{(\beta)}_{t,i-1}N_t^{\varepsilon}G_{t,i}\mid\mathcal H_{t,i-1}]\\
&=M^{(\beta)}_{t,i-1}\E[N_t^{\varepsilon}G_{t,i}\mid\mathcal H_{t,i-1}]\\
&\le M^{(\beta)}_{t,i-1}N_t^{\varepsilon}\\
&=X_{i-1}.
\end{align*}
Taking $i=m+1$ gives the required conditional inequality. Together with the nonnegativity, measurability, and integrability established above, this proves that $(X_m)_{m=0}^{n}$ is a nonnegative supermartingale. Since $M^{(\beta)}_{t,0}=1$, its initial value is $X_0=N_t^{\varepsilon}$.

For a stopping time $\tau_t\le n$, the actual stopped process $X_{m\wedge\tau_t}$ is also a supermartingale: its next increment is $\ind\{\tau_t>m\}(X_{m+1}-X_m)$, with a predictable indicator and nonpositive conditional mean. Iterating its conditional inequality from 0 to $n$ yields
\[
\E[X_{n\wedge\tau_t}\mid\mathcal H_{t,0}]\le X_0=N_t^{\varepsilon},
\qquad
N_t^{\varepsilon}\E[M^{(\beta)}_{t,\tau_t}\mid\Fpre_t]\le N_t^{\varepsilon}.
\]
On $H_t^0$, we have $N_t^{\varepsilon}=1$. Therefore, for every stopping time $\tau_t\le n$,
\[
\E[M^{(\beta)}_{t,\tau_t}\mid\Fpre_t]\le1
\qquad\text{almost surely on }H_t^0.
\]
This is the stopping extension of Theorem~\ref{thm:evidence}. In particular, taking $\tau_t=n$ gives
\[
\E[M^{(\beta)}_{t,n}\mid\Fpre_t]\le1
\qquad\text{almost surely on }H_t^0,
\]
which is Equation~\eqref{eq:conditional-evalidity}.
\end{proof}

\subsection{Stopping times}

\begin{proposition}
\label{prop:crossing}
Suppose $\alpha_t\in(0,1)$ is $\Fpre_t$-measurable and set
\[
\tau_t
=\inf\left\{m\in\{0,\ldots,n\}:
M^{(\beta)}_{t,m}\ge\frac1{\alpha_t}\right\}\wedge n.
\]
Then $\tau_t$ is a stopping time bounded by $n$, and
\[
\ind\{M^{(\beta)}_{t,\tau_t}\ge1/\alpha_t\}
=\ind\left\{\max_{0\le m\le n}M^{(\beta)}_{t,m}\ge1/\alpha_t\right\}.
\]
\end{proposition}

\begin{proof}
For $0\le m<n$,
\[
\{\tau_t\le m\}
=\bigcup_{j=0}^m\{M^{(\beta)}_{t,j}\ge1/\alpha_t\}
\in\mathcal H_{t,m},
\]
because $M^{(\beta)}_{t,j}$ is $\mathcal H_{t,j}$-measurable, $\alpha_t$ is $\mathcal H_{t,0}=\Fpre_t$-measurable, and $\mathcal H_{t,0}\subseteq\mathcal H_{t,j}\subseteq\mathcal H_{t,m}$ for every $j\le m$.

By definition, $\tau_t\le n$ for every sample outcome. Therefore, when $m\ge n$, we have $\tau_t\le n\le m$ for every outcome, so
\[
\{\tau_t\le m\}=\Omega,
\]
where $\Omega$ is the set of all possible sample outcomes. In particular, $\{\tau_t\le n\}=\Omega\in\mathcal H_{t,n}$. Together with the case $m<n$ above, this proves that $\tau_t$ is a stopping time bounded by $n$.

If a crossing occurs, the stopped value is the first crossing value; otherwise the process stops at $n$ below threshold.
\end{proof}

\subsection{Mixtures}

\paragraph{Betting fractions.}
Betting interprets an evidence process as capital accumulated from an initial value of one \citep{waudbysmith2024estimating}. In Equation~\eqref{eq:reuse-bet-y}, the amount bet on observation $i$ is $\eta_{t,i}M^{(\beta)}_{t,i-1}$, where $M^{(\beta)}_{t,i-1}$ is the current capital and $\eta_{t,i}\in[0,1)$ specifies the proportion of that capital put at risk. The new capital equals the current capital plus the gain or loss from this bet:
\begin{align*}
M^{(\beta)}_{t,i}
&=M^{(\beta)}_{t,i-1}
+\eta_{t,i}M^{(\beta)}_{t,i-1}
\frac{Y_{t,i}-u_{t,i}^{(\beta)}}{u_{t,i}^{(\beta)}}\\
&=M^{(\beta)}_{t,i-1}
\left[(1-\eta_{t,i})
+\eta_{t,i}\frac{Y_{t,i}}{u_{t,i}^{(\beta)}}\right].
\end{align*}
The uncommitted capital is retained, and the amount bet is multiplied by the return factor $Y_{t,i}/u_{t,i}^{(\beta)}$. Thus $\eta_{t,i}=0$ leaves the evidence unchanged; if $Y_{t,i}=0$, the update loses exactly the fraction $\eta_{t,i}$ of current capital. Larger betting fractions increase both the gain when $Y_{t,i}>u_{t,i}^{(\beta)}$ and the loss when $Y_{t,i}<u_{t,i}^{(\beta)}$. The amount of each bet is calculated using the current capital. A betting strategy specifies how to choose these fractions from the available history: each $\eta_{t,i}$ must be $\mathcal H_{t,i-1}$-measurable, so it is chosen before observing $Y_{t,i}$.

\paragraph{Combining betting strategies.}
Let $L\ge1$ be a fixed integer denoting the number of betting strategies included in the mixture. For each $\ell\in\{1,\ldots,L\}$, strategy $\ell$ uses $\mathcal H_{t,i-1}$-measurable fractions $\eta_{t,i}^{(\ell)}\in[0,1)$ and produces
\[
M_{t,0}^{(\beta,\ell)}=1,
\qquad
M_{t,m}^{(\beta,\ell)}
=\prod_{i=1}^{m}
\left[1+\eta_{t,i}^{(\ell)}
\frac{Y_{t,i}-u_{t,i}^{(\beta)}}{u_{t,i}^{(\beta)}}\right].
\]
All $L$ processes use the same validation observations and corrected boundaries, with potentially different betting fractions. A mixture combines their evidence by taking a weighted average, allowing several betting strategies to contribute to the test \citep{waudbysmith2024estimating}. The weight $\pi_{t,\ell}$ is the fraction of initial capital assigned to strategy $\ell$; the weights are nonnegative, sum to one, are $\Fpre_t$-measurable, and remain fixed throughout the round. Convex averaging preserves e-value validity even when the component e-values depend on the same observations \citep{vovk2021evalues}. The following corollary establishes the corresponding conditional validity for these sequential processes.

\begin{corollary}
\label{cor:mixture}
Let $M_{t,m}^{(\beta,1)},\ldots,M_{t,m}^{(\beta,L)}$ use predictable betting sequences and let nonnegative weights $\pi_{t,1},\ldots,\pi_{t,L}$ be $\Fpre_t$-measurable with $\sum_{\ell=1}^L\pi_{t,\ell}=1$. Then $M_{t,m}^{(\beta,\mathrm{mix})}=\sum_{\ell=1}^L\pi_{t,\ell}M_{t,m}^{(\beta,\ell)}$ is a nonnegative supermartingale under the corrected null, and $M_{t,\tau_t}^{(\beta,\mathrm{mix})}$ is a conditionally valid e-value at every bounded stopping time.
\end{corollary}

\begin{proof}[Proof of Corollary~\ref{cor:mixture}]
For each $\ell$, conditional supermartingale validity gives
\[
\E[M_{t,i}^{(\beta,\ell)}\mid\mathcal H_{t,i-1}]
\le M_{t,i-1}^{(\beta,\ell)}.
\]
The weights are nonnegative, sum to one, and are known at $\mathcal H_{t,0}=\Fpre_t$. Hence
\begin{align*}
\E[M_{t,i}^{(\beta,\mathrm{mix})}\mid\mathcal H_{t,i-1}]
&=\sum_{\ell=1}^L\pi_{t,\ell}
\E[M_{t,i}^{(\beta,\ell)}\mid\mathcal H_{t,i-1}]\\
&\le\sum_{\ell=1}^L\pi_{t,\ell}M_{t,i-1}^{(\beta,\ell)}
=M_{t,i-1}^{(\beta,\mathrm{mix})}.
\end{align*}
Nonnegativity is immediate, and bounded optional stopping gives the stopped e-value claim.
\end{proof}

\section{Proof of False Evolution Rate Control}
\label{app:fer-proof}

\begin{lemma}
\label{lem:egate}
If $\alpha_t\in[0,1]$ is $\Fpre_t$-measurable and $E_t$ is conditionally valid under $H_t^0$, then
\[
\E[N_t^{\varepsilon}A_t\mid\Fpre_t]\le N_t^{\varepsilon}\alpha_t,
\]
where $\alpha_t=0$ is interpreted as $A_t=0$.
\end{lemma}

\begin{proof}
When $\alpha_t>0$, the indicator inequality
$\ind\{E_t\ge1/\alpha_t\}\le\alpha_tE_t$ holds. When $\alpha_t=0$, both $A_t$ and $\alpha_tE_t$ are zero. Thus $A_t\le\alpha_tE_t$ in all cases. Since $N_t^{\varepsilon}$ and $\alpha_t$ are $\Fpre_t$-measurable,
\[
\E[N_t^{\varepsilon}A_t\mid\Fpre_t]
\le N_t^{\varepsilon}\alpha_t\E[E_t\mid\Fpre_t]
\le N_t^{\varepsilon}\alpha_t.
\]
\end{proof}

\subsection{Deterministic horizon control in Theorem~\ref{thm:fer}}

\begin{proof}
For $\alpha_t>0$, Algorithm~\ref{alg:TRUST} uses $M^{(\beta)}_{t,\tau_t}$ as its evidence, with the stopping time $\tau_t\le n$ defined in Proposition~\ref{prop:crossing}. The stopping extension of Theorem~\ref{thm:evidence} ensures that this evidence is conditionally valid on $H_t^0$. Therefore, Lemma~\ref{lem:egate} gives
\[
\E[N_t^{\varepsilon}A_t\mid\Fpre_t]\le N_t^{\varepsilon}\alpha_t.
\]
When $\alpha_t=0$, the algorithm sets $A_t=0$, so the same inequality holds.

We first show that each testing level $\alpha_t$ is $\Fpre_t$-measurable and that the total budget spent is at most $q$. At the first round, $B_1=q$ and $R_0=0$. The baseline $\rho_t^0=1/(T-t+1)$ is deterministic for $1\le t\le T$, and $w_t$ is selected using only $\Fpre_t$. Thus \eqref{eq:spending-fraction} and~\eqref{eq:testing-level} make $\rho_t$, $\alpha_t$, and
\[
c_t=\frac{\alpha_t}{R_{t-1}+1}
=\min\left\{\rho_tB_t,\frac{\alpha_{\max}}{R_{t-1}+1}\right\}
\]
$\Fpre_t$-measurable. Hence the current testing level is fixed before validation. Moreover, $0\le\rho_t\le1$ implies
\[
0\le c_t\le\rho_tB_t\le B_t,
\]
so $B_{t+1}=B_t-c_t$ remains nonnegative. For any $m\le T$, summing $c_t=B_t-B_{t+1}$ over $t=1,\ldots,m$ gives
\[
\sum_{t=1}^m\frac{\alpha_t}{R_{t-1}+1}
=\sum_{t=1}^m(B_t-B_{t+1})
=q-B_{m+1}\le q.
\]

If $A_t=1$ and $t\le T$, the total number of acceptances by horizon $T$ is at least the earlier count plus the current acceptance:
\[
R_T\ge R_{t-1}+1.
\]
If $A_t=0$, the current false-acceptance numerator vanishes. Hence, we have 
\[
\frac{N_t^{\varepsilon}A_t}{R_T\vee1}
\le\frac{N_t^{\varepsilon}A_t}{R_{t-1}+1}.
\]
Summing and taking expectations yields
\begin{align*}
\FER_\varepsilon(T)
&=\E\left[\frac{\sum_{t=1}^TN_t^{\varepsilon}A_t}{R_T\vee1}\right]\\
&\le\sum_{t=1}^T\E\left[\frac{N_t^{\varepsilon}A_t}{R_{t-1}+1}\right]\\
&=\sum_{t=1}^T\E\left[
\frac{\E[N_t^{\varepsilon}A_t\mid\Fpre_t]}{R_{t-1}+1}
\right]\\
&\le\sum_{t=1}^T\E\left[
\frac{N_t^{\varepsilon}\alpha_t}{R_{t-1}+1}
\right]\\
&\le\E\left[
\sum_{t=1}^T\frac{\alpha_t}{R_{t-1}+1}
\right]
\le q.
\end{align*}
The third line uses that $R_{t-1}$ is $\Fpre_t$-measurable, and the fourth uses Lemma~\ref{lem:egate}.
\end{proof}

\subsection{Outer  stopping extension of Theorem~\ref{thm:fer}}
\label{app:outer-stopping}

\begin{proof}
We extend the run beyond its planned horizon $T$ by setting $A_t=\alpha_t=c_t=0$ for $t>T$ and keeping the history and account balance fixed.
Let $\tau$ be an almost surely finite stopping time for $(\F_t)_{t\ge0}$. The stopped counts admit nonnegative series representations
\[
F_\tau^\varepsilon
=\sum_{t=1}^{\infty}\ind\{\tau\ge t\}N_t^{\varepsilon}A_t,
\qquad
R_\tau
=\sum_{t=1}^{\infty}\ind\{\tau\ge t\}A_t.
\]
Only finitely many summands are nonzero on almost every path. If $\tau\ge t$ and $A_t=1$, then $R_\tau\ge R_{t-1}+1$. Consequently,
\[
\frac{F_\tau^\varepsilon}{R_\tau\vee1}
\le
\sum_{t=1}^{\infty}
\ind\{\tau\ge t\}
\frac{N_t^{\varepsilon}A_t}{R_{t-1}+1}.
\]
The stopping-time property implies
$\{\tau\ge t\}=\{\tau\le t-1\}^c\in\F_{t-1}\subseteq\Fpre_t$. Therefore the survival indicator is predictable, and Lemma~\ref{lem:egate} gives
\[
\E\left[
\ind\{\tau\ge t\}\frac{N_t^{\varepsilon}A_t}{R_{t-1}+1}
\middle|\Fpre_t
\right]
\le
\ind\{\tau\ge t\}
\frac{\alpha_t}{R_{t-1}+1}.
\]
All summands are nonnegative, so Tonelli's theorem and the tower property yield
\begin{align*}
\E\left[\frac{F_\tau^\varepsilon}{R_\tau\vee1}\right]
&\le
\E\left[
\sum_{t=1}^{\infty}
\ind\{\tau\ge t\}\frac{\alpha_t}{R_{t-1}+1}
\right]\\
&\le
\E\left[
\sum_{t=1}^{\infty}\frac{\alpha_t}{R_{t-1}+1}
\right]
\le q.
\end{align*}
Since $c_t=0$ for $t>T$, the last inequality follows from $\sum_{t=1}^{T}c_t=q-B_{T+1}\le q$. This extension is stronger than guaranteeing each fixed deterministic horizon separately: the stopping time itself can depend on the observed recursive self-improvement history.
\end{proof}

\section{Proof of True Evolution Rate Convergence}

\subsection{Proof of Theorem~\ref{thm:ter}}
\label{app:consistency-proof}

\begin{proof}
Fix $T<\infty$ and a sequence of runs satisfying Assumption~\ref{ass:resource-consistency}. We suppress dependence of the candidates and histories on $(N,n)$. All statements for sufficiently large $(N,n)$ are along this sequence, and all conditional inequalities below hold almost surely.

\paragraph{Step 1: Bound the testing levels.}
Equation~\eqref{eq:budget-update} gives
\[
c_t=\frac{\alpha_t}{R_{t-1}+1}
=\min\left\{\rho_tB_t,\frac{\alpha_{\max}}{R_{t-1}+1}\right\}
\le\rho_{\max}B_t.
\]
Consequently, $B_{t+1}=B_t-c_t\ge(1-\rho_{\max})B_t$. Starting from $B_1=q$ and iterating yields
\[
B_t\ge q(1-\rho_{\max})^{t-1}
\ge q(1-\rho_{\max})^{T-1},
\qquad t\le T.
\]
Since $\rho_t\ge\rho_{\min}$ and $R_{t-1}+1\ge1$, we obtain the deterministic lower bound
\begin{equation}
\alpha_t\ge\alpha_*
:=\min\left\{\alpha_{\max},
\rho_{\min}q(1-\rho_{\max})^{T-1}\right\}>0.
\label{eq:resource-alpha-lower}
\end{equation}
Write $\Lambda=\log(1/\alpha_*)$. Both $\alpha_*$ and $\Lambda$ are independent of $(N,n)$.

\paragraph{Step 2: Bound exposure and the conditional mean under improvement.}
At the beginning of round $t$, at most $(t-1)n$ distinct validation indices have been exposed. An index drawn earlier in the same round has been removed from $\mathcal R_{t,i-1}$, so it cannot contribute to $\mathcal I_{t,i-1}\cap\mathcal R_{t,i-1}$. Thus, for every draw,
\begin{equation}
0\le\zeta_{t,i}
\le\bar\zeta_{N,n}
:=\min\left\{1,\frac{(T-1)n}{N-n+1}\right\}.
\label{eq:resource-exposure-bound}
\end{equation}
Because $T$ is fixed and $n/N\to0$, we have $\bar\zeta_{N,n}\to0$.

Let $L_{t,i}$ denote the $\mathcal H_{t,i-1}$-measurable indicator that draw $i$ is made. On $\{\Delta_t>0,L_{t,i}=1\}$, Equation~\eqref{eq:unexposed-mean} and $Y_{t,i}Q_{t,i}\ge0$ give
\begin{align*}
\E[Y_{t,i}\mid\mathcal H_{t,i-1}]
&\ge\E[Y_{t,i}(1-Q_{t,i})\mid\mathcal H_{t,i-1}]\\
&=(1-\zeta_{t,i})\frac{1+\Delta_t}{2}\\
&\ge\frac{1-\bar\zeta_{N,n}}{2}.
\end{align*}
Also,
\[
u_{t,i}^{(\beta)}=p_\varepsilon+\beta\zeta_{t,i}
\le p_\varepsilon+\beta\bar\zeta_{N,n}.
\]

\paragraph{Step 3: Positive log-growth bound.}
For $z\in[0,1]$ and $\eta\in[\eta_-,\eta_+]$, define
\[
\phi(z,\eta)
:=\frac{1+z}{2}\log(1-\eta)
+\frac{1-z}{2}\log\left(1-\eta+
\frac{\eta}{p_\varepsilon+\beta z}\right).
\]
On a draw, the log increment is
\[
V_{t,i}:=\log\frac{M^{(\beta)}_{t,i}}{M^{(\beta)}_{t,i-1}}
=\log\left(1-\eta_{t,i}+
\frac{\eta_{t,i}Y_{t,i}}{u_{t,i}^{(\beta)}}\right).
\]
Set $z=\bar\zeta_{N,n}$ and $\bar u=p_\varepsilon+\beta z$. Conditional on $\mathcal H_{t,i-1}$, $\eta_{t,i}$ is fixed. Since $u_{t,i}^{(\beta)}\le\bar u$ and $Y_{t,i}\ge0$,
\[
V_{t,i}\ge\log\left(1-\eta_{t,i}+
\frac{\eta_{t,i}Y_{t,i}}{\bar u}\right).
\]
The function $\log(1-\eta+\eta y/\bar u)$ is concave on $[0,1]$, so we have:
\[
\log\left(1-\eta+\frac{\eta y}{\bar u}\right)
\ge(1-y)\log(1-\eta)
+y\log\left(1-\eta+\frac{\eta}{\bar u}\right).
\]
The right-hand side is increasing in $y$. 
Applying this inequality with $y=Y_{t,i}$ and
$\eta=\eta_{t,i}$, and then taking conditional expectations, gives
\begin{align*}
\E[V_{t,i}\mid\mathcal H_{t,i-1}]
\ge{}&
\left(1-m_{t,i}\right)\log(1-\eta_{t,i})+
m_{t,i}
\log\left(
1-\eta_{t,i}
+\frac{\eta_{t,i}}{\bar u}
\right),
\end{align*}
where
\(m_{t,i}:=\E[Y_{t,i}\mid\mathcal H_{t,i-1}].\) By Step~2, on $\{\Delta_t>0,\ L_{t,i}=1\}$,
\[
m_{t,i}
\ge
\frac{1-\bar\zeta_{N,n}}{2}.
\]
Moreover,
\[
\log\left(
1-\eta_{t,i}
+\frac{\eta_{t,i}}{\bar u}
\right)
>
\log(1-\eta_{t,i}),
\]
so the preceding lower bound is increasing in $m_{t,i}$. Therefore,
\begin{align*}
\E[V_{t,i}\mid\mathcal H_{t,i-1}]
\ge{}&
\frac{1+\bar\zeta_{N,n}}{2}
\log(1-\eta_{t,i})\\
&+
\frac{1-\bar\zeta_{N,n}}{2}
\log\left(
1-\eta_{t,i}
+
\frac{\eta_{t,i}}
{p_\varepsilon+\beta\bar\zeta_{N,n}}
\right)\\
={}&
\phi(\bar\zeta_{N,n},\eta_{t,i}).
\end{align*}

For fixed $z$, $\phi(z,\eta)$ is concave in $\eta$, hence its minimum on $[\eta_-,\eta_+]$ is attained at an endpoint. Define
\begin{equation}
g_{N,n}
:=\min\{\phi(\bar\zeta_{N,n},\eta_-),
          \phi(\bar\zeta_{N,n},\eta_+)\}.
\label{eq:resource-log-growth}
\end{equation}
It follows that $\E[V_{t,i}\mid\mathcal H_{t,i-1}]\ge g_{N,n}$ on every improving draw.

Continuity and $\bar\zeta_{N,n}\to0$ imply
\[
g_{N,n}\longrightarrow
 g_0:=\min\{\phi(0,\eta_-),\phi(0,\eta_+)\}.
\]
To verify that $g_0>0$, note that
\[
\phi(0,\eta)
=\frac12\log\left[(1-\eta)
\left(1-\eta+\frac{\eta}{p_\varepsilon}\right)\right]
\]
and
\begin{equation}
\label{eq:etadiff}
(1-\eta)\left(1-\eta+\frac{\eta}{p_\varepsilon}\right)-1
=\frac{\eta}{p_\varepsilon}
\left[\varepsilon-\eta(1-p_\varepsilon)\right].
\end{equation}
Assumption~\ref{ass:resource-consistency}(c) gives
\[
0<\eta_-\le\eta\le\eta_+
<\frac{2\varepsilon}{1+\varepsilon}
=\frac{\varepsilon}{1-p_\varepsilon}.
\]
The right-hand side of Equation~\eqref{eq:etadiff} is therefore strictly positive, so both endpoint values defining $g_0$ are positive. Since $\bar\zeta_{N,n}\to0$ and $\phi(z,\eta)$ is continuous in $z$,
\[
\phi(\bar\zeta_{N,n},\eta_-)\to\phi(0,\eta_-),
\qquad
\phi(\bar\zeta_{N,n},\eta_+)\to\phi(0,\eta_+).
\]
Therefore,
\[
g_{N,n}
=
\min\{\phi(\bar\zeta_{N,n},\eta_-),
      \phi(\bar\zeta_{N,n},\eta_+)\}
\to
g_0,
\]
where \(g_0 =\min\{\phi(0,\eta_-),\phi(0,\eta_+)\}>0.\)
Hence, for all sufficiently large $N$ and $n$,
\[
|g_{N,n}-g_0|<\frac{g_0}{2},
\]
which implies
\[
g_{N,n}\ge \frac{g_0}{2}>0.
\]

The log increments also lie in a fixed bounded interval. Since $u_{t,i}^{(\beta)}\ge p_\varepsilon$, $Y_{t,i}\in[0,1]$, and $\eta_{t,i}\le\eta_+$ on draws,
\[
\log(1-\eta_+)
\le V_{t,i}
\le\log\left(1-\eta_++\frac{\eta_+}{p_\varepsilon}\right).
\]
After stopping, $V_{t,i}=0$, which lies in the same interval. Its width is the fixed constant
\begin{equation}
h:=\log\left(
\frac{1-\eta_++\eta_+/p_\varepsilon}{1-\eta_+}
\right)>0.
\label{eq:resource-log-width}
\end{equation}
Thus the log increments are integrable, and the conditions of Appendix~\ref{app:log-growth} hold with $g_t=g_{N,n}$ and $h_t=h$ for all sufficiently large runs.

\paragraph{Step 4: Bound conditional miss probabilities.}
Applying Lemma~\ref{lem:conditional-miss} with $g_t=g_{N,n}$ and $h_t=h$ gives, on $\{\Delta_t>0\}$,
\begin{equation}
\begin{aligned}
\Prb(A_t=0\mid\Fpre_t)
&\le\exp\left[-\frac{2n}{h^2}
\left(g_{N,n}-\frac{\log(1/\alpha_t)}{n}\right)_+^2\right]\\
&\le\exp\left[-\frac{2n}{h^2}
\left(g_{N,n}-\frac{\Lambda}{n}\right)_+^2\right].
\end{aligned}
\label{eq:resource-finite-miss-bound}
\end{equation}
The second inequality follows from $\alpha_t\ge\alpha_*$ and hence $\log(1/\alpha_t)\le\Lambda$.

For sufficiently large $N$ and $n$, we have $g_{N,n}\ge g_0/2$ and $n\ge4\Lambda/g_0$. Therefore,
\[
g_{N,n}-\frac{\Lambda}{n}\ge\frac{g_0}{4}.
\]
Set
\begin{equation}
\kappa:=\frac{g_0^2}{8h^2}>0.
\label{eq:resource-kappa}
\end{equation}
This constant is independent of $(N,n)$, and Equation~\eqref{eq:resource-finite-miss-bound} yields
\begin{equation}
\Prb(A_t=0\mid\Fpre_t)
\le e^{-\kappa n}
\qquad\text{on }\{\Delta_t>0\},
\quad t=1,\ldots,T.
\label{eq:derived-miss-bound}
\end{equation}
In particular, the conditional miss bound is a consequence of the sampling, allocation, and betting rules, rather than an assumed property of the gate.

\paragraph{Step 5: TER consistency.}
Let $L_T^+=\sum_{t=1}^TI_t(1-A_t)=U_T-S_T$ denote the number of improving candidates missed. Since $I_t$ is $\Fpre_t$-measurable, Equation~\eqref{eq:derived-miss-bound} and the tower property give
\begin{align*}
\E[L_T^+]
&=\sum_{t=1}^T\E\left[I_t\Prb(A_t=0\mid\Fpre_t)\right]\\
&\le e^{-\kappa n}\sum_{t=1}^T\E[I_t]\\
&\le Te^{-\kappa n}.
\end{align*}
The following identity holds on every trajectory:
\[
1-\frac{S_T}{U_T\vee1}
=\ind\{U_T=0\}+\frac{U_T-S_T}{U_T\vee1}.
\]
Indeed, when $U_T=0$, also $S_T=0$, and both sides equal one. When $U_T>0$, the indicator vanishes and the equality follows from $U_T\vee1=U_T$. Taking expectations, using $U_T\vee1\ge1$, and then applying Assumption~\ref{ass:resource-consistency}(a), we obtain
\begin{align*}
1-\TER(T)
&=\Prb(U_T=0)+\E\left[\frac{L_T^+}{U_T\vee1}\right]\\
&\le\Prb(U_T=0)+\E[L_T^+]\\
&\le r_{N,n}+Te^{-\kappa n}.
\end{align*}
Rearranging proves Equation~\eqref{eq:ter-consistency}. Because $T$ is fixed, $r_{N,n}\to0$, and $n\to\infty$, the right-hand side tends to zero. Since $0\le\TER(T)\le1$, we conclude that $\TER(T)\to1$. No independence between the future candidate count $U_T$ and current acceptance decisions is used.
\end{proof}

\subsection{An explicit conditional miss bound}
\label{app:log-growth}

The following lemma provides the conditional miss bound used to derive Equation~\eqref{eq:resource-finite-miss-bound} in Step~4 of the proof of Theorem~\ref{thm:ter}. Since $\eta_{t,i}<1$, every factor in Equation~\eqref{eq:reuse-bet-y} is at least $1-\eta_{t,i}>0$, so $M^{(\beta)}_{t,i}>0$ and its log increments are well defined. For each round with $\alpha_t>0$, define
\[
V_{t,i}
=
\log\frac{M^{(\beta)}_{t,i}}{M^{(\beta)}_{t,i-1}},
\]
and let $L_{t,i}$ be the $\mathcal H_{t,i-1}$-measurable indicator that validation draw $i$ is made.

\begin{lemma}[Conditional miss bound]
\label{lem:conditional-miss}
Suppose there exist finite $\Fpre_t$-measurable $g_t,h_t>0$ and finite $\mathcal H_{t,i-1}$-measurable interval endpoints $a_{t,i},b_{t,i}$ such that, on $\{\Delta_t>0,\ L_{t,i}=1\}$, for every $i=1,\ldots,n$,
\begin{equation}
\E[V_{t,i}\mid\mathcal H_{t,i-1}]\ge g_t,
\qquad
a_{t,i}\le V_{t,i}\le b_{t,i},
\qquad
b_{t,i}-a_{t,i}\le h_t.
\label{eq:log-growth-conditions}
\end{equation}
Then
\[
\Prb(A_t=0\mid\Fpre_t)\le\epsilon_t
\qquad\text{on }\{\Delta_t>0\},
\]
where
\begin{equation}
\epsilon_t=
\begin{cases}
\displaystyle
\exp\left[-\frac{2n}{h_t^2}
\left(g_t-\frac{\log(1/\alpha_t)}{n}\right)_+^2\right],
& \alpha_t>0,\\
1, & \alpha_t=0.
\end{cases}
\label{eq:conditional-miss-bound}
\end{equation}
\end{lemma}

\begin{proof}
If $\alpha_t=0$, then $A_t=0$ and $\epsilon_t=1$, so the miss bound holds.

Work on the $\Fpre_t$-measurable event $\{\Delta_t>0,\ \alpha_t>0\}$ throughout the following calculation. Define
\[
\mu_{t,i}=\E[V_{t,i}\mid\mathcal H_{t,i-1}],
\qquad
Z_{t,i}=V_{t,i}-\mu_{t,i}.
\]
On $\{L_{t,i}=1\}$, Equation~\eqref{eq:log-growth-conditions} gives $\mu_{t,i}\ge g_t$, and subtracting $\mu_{t,i}$ from the interval endpoints preserves their width, which is at most $h_t$. On $\{L_{t,i}=0\}$, the evidence is unchanged, so $V_{t,i}=0$. Since this event is $\mathcal H_{t,i-1}$-measurable, $\mu_{t,i}=Z_{t,i}=0$ there as well. The log increments satisfy $V_{t,i}\le\log(1/p_\varepsilon)$ and have nonnegative conditional means on the event under consideration, so their negative parts have conditional expectations at most $\log(1/p_\varepsilon)$. They are therefore integrable when restricted to this event, and $\E[Z_{t,i}\mid\mathcal H_{t,i-1}]=0$ there.

For any finite nonnegative $\Fpre_t$-measurable $\lambda$, the conditional Hoeffding lemma gives
\[
\E[e^{-\lambda Z_{t,i}}\mid\mathcal H_{t,i-1}]
\le\exp(\lambda^2h_t^2/8).
\]
Both $\lambda$ and $h_t$ are fixed when conditioning on $\Fpre_t$. Dividing each factor by $\exp(\lambda^2h_t^2/8)$ and iterating conditional expectations from $\mathcal H_{t,0}=\Fpre_t$ yields
\[
\E\left[
\exp\left\{-\lambda\sum_{i=1}^{n}Z_{t,i}
-\frac{n\lambda^2h_t^2}{8}\right\}
\middle|\Fpre_t
\right]\le1.
\]
If $A_t=0$, all $n$ validation draws are made and the evidence never reaches $1/\alpha_t$. Therefore,
\[
\sum_{i=1}^{n}V_{t,i}
=\log M^{(\beta)}_{t,n}<\log(1/\alpha_t).
\]
On this miss event, $L_{t,i}=1$ for every $i=1,\ldots,n$. Equation~\eqref{eq:log-growth-conditions} then gives $\mu_{t,i}\ge g_t$ at every draw, so
\[
\sum_{i=1}^{n}\mu_{t,i}\ge ng_t.
\]
Define $d_t=g_t-\log(1/\alpha_t)/n$. Subtracting the conditional means from the realized log increments gives, on $\{A_t=0\}$,
\begin{align*}
\sum_{i=1}^{n}Z_{t,i}
&=\sum_{i=1}^{n}V_{t,i}-\sum_{i=1}^{n}\mu_{t,i}\\
&<\log(1/\alpha_t)-ng_t\\
&=-n\left(g_t-\frac{\log(1/\alpha_t)}{n}\right)\\
&=-nd_t.
\end{align*}
Thus every outcome with $A_t=0$ satisfies this inequality, which proves the event inclusion
\[
\{A_t=0\}\subseteq
\left\{\sum_{i=1}^{n}Z_{t,i}<-nd_t\right\}.
\]
If $d_t\le0$, the formula gives $\epsilon_t=1$, and the miss probability is at most one. On $\{d_t>0\}$, take a finite positive $\Fpre_t$-measurable $\lambda$. The event inclusion above implies that, whenever $A_t=0$,
\[
\exp\left\{-\lambda\sum_{i=1}^{n}Z_{t,i}
-\frac{n\lambda^2h_t^2}{8}\right\}
>
\exp\left(\lambda nd_t-\frac{n\lambda^2h_t^2}{8}\right).
\]
The random variable on the left is nonnegative. The threshold on the right is positive and $\Fpre_t$-measurable, since $\lambda$, $d_t$, and $h_t$ are all $\Fpre_t$-measurable. Conditional Markov's inequality bounds the probability of exceeding this threshold by the conditional expectation divided by the threshold. Therefore,
\begin{align*}
\Prb(A_t=0\mid\Fpre_t)
&\le
\frac{
\displaystyle\E\left[
\exp\left\{-\lambda\sum_{i=1}^{n}Z_{t,i}
-\frac{n\lambda^2h_t^2}{8}\right\}
\middle|\Fpre_t
\right]
}{
\exp\left(\lambda nd_t-\frac{n\lambda^2h_t^2}{8}\right)
}\\
&\le
\frac{1}{\exp\left(\lambda nd_t-\frac{n\lambda^2h_t^2}{8}\right)}\\
&=\exp\left(-\lambda nd_t+\frac{n\lambda^2h_t^2}{8}\right).
\end{align*}
The second inequality uses the conditional expectation bound of one established above by iterating the conditional Hoeffding bound.
The choice $\lambda=4(d_t)_+/h_t^2$ is $\Fpre_t$-measurable and gives $\exp(-2nd_t^2/h_t^2)$ when $d_t>0$. Combining these cases with $\alpha_t=0$ proves Equation~\eqref{eq:conditional-miss-bound}.

\end{proof}

\subsection{Extension of the power bound to mixtures}
\label{app:mixture-power}

Consider the convex mixture in Corollary~\ref{cor:mixture},
\[
M_{t,m}^{(\beta,\mathrm{mix})}
=\sum_{\ell=1}^{L}\pi_{t,\ell}M_{t,m}^{(\beta,\ell)},
\qquad
\pi_{t,\ell}\ge0,
\quad\sum_{\ell=1}^{L}\pi_{t,\ell}=1,
\]
where the weights are $\Fpre_t$-measurable and remain fixed during validation. Suppose at least one component $\ell^\star$ has predictable fractions satisfying Assumption~\ref{ass:resource-consistency}(c) on draws and weights $\pi_{t,\ell^\star}\ge\pi_*>0$ for every round and run, with $\pi_*$ independent of $(N,n)$. Other components need not satisfy this power condition, but they remain nonnegative processes constructed by Equation~\eqref{eq:reuse-bet-y} with predictable fractions in $[0,1)$.

\begin{proof}
Nonnegativity gives $M_{t,m}^{(\beta,\mathrm{mix})}\ge\pi_*M_{t,m}^{(\beta,\ell^\star)}$. If the mixture gate misses an improving candidate, all $n$ draws are made and
\[
M_{t,n}^{(\beta,\mathrm{mix})}<1/\alpha_t,
\qquad
M_{t,n}^{(\beta,\ell^\star)}<1/(\pi_*\alpha_t).
\]
Steps 2--3 of the proof of Theorem~\ref{thm:ter} apply to the selected component, with its increments held at zero after the mixture gate stops. The conditional concentration argument in Appendix~\ref{app:log-growth} therefore gives
\[
\Prb(A_t=0\mid\Fpre_t)
\le\exp\left[-\frac{2n}{h^2}
\left(g_{N,n}-\frac{\log(1/(\pi_*\alpha_t))}{n}\right)_+^2\right]
\]
on $\{\Delta_t>0\}$. The additional threshold penalty is at most the fixed constant $\log(1/\pi_*)$. Replacing $\Lambda$ by $\Lambda+\log(1/\pi_*)$ in Step 4 thus gives the same $e^{-\kappa n}$ bound for all sufficiently large runs. Step 5 then proves the same TER conclusion. Under Assumption~\ref{ass:exposure-bias}, the mixture remains conditionally valid by Corollary~\ref{cor:mixture}, so the proof of Theorem~\ref{thm:utility} below also applies. Only the designated component is required to satisfy the power condition.
\end{proof}

\section{Proof of the Gain Bound}
\label{app:gain-proof}

Define the cumulative loss from accepted updates by
\[
D_T^-=\sum_{t=1}^T A_t(-\Delta_t)_+.
\]

\begin{proof}[Proof of Theorem~\ref{thm:utility}]
Fix a finite horizon $T\ge1$ and the resource sequence and algorithm configuration in Appendix~\ref{app:consistency-proof}. Take $\kappa>0$ as in Equation~\eqref{eq:resource-kappa} and consider a sufficiently large run for Equation~\eqref{eq:derived-miss-bound} to hold. Throughout, the candidates, effects, and acceptance decisions refer to the actual adaptive trajectory.

\paragraph{Step 1: Bound cumulative accepted loss.}
Under Assumption~\ref{ass:exposure-bias}, Theorem~\ref{thm:evidence} and Proposition~\ref{prop:crossing} provide conditionally valid evidence at the gate's stopping time. Lemma~\ref{lem:egate} therefore gives
\[
\E[N_t^{\varepsilon}A_t\mid\Fpre_t]\le N_t^{\varepsilon}\alpha_t.
\]
Let $W_t=(-\Delta_t-\varepsilon)_+$. Since $\Delta_t$ is $\Fpre_t$-measurable and lies in $[-1,1]$, $W_t$ is bounded, nonnegative, and $\Fpre_t$-measurable. Moreover, $W_t>0$ implies $\Delta_t<-\varepsilon$, so $W_t=N_t^{\varepsilon}W_t$. Thus
\[
\E[A_tW_t]
=\E\left[W_t\E[N_t^{\varepsilon}A_t\mid\Fpre_t]\right]
\le\E[\alpha_tW_t].
\]
The identity
\[
(-\Delta_t)_+
=\min\{(-\Delta_t)_+,\varepsilon\}+(-\Delta_t-\varepsilon)_+
\]
splits each accepted loss into the part up to $\varepsilon$ and the excess. Multiplying by $A_t$, summing, and taking expectations gives
\begin{align*}
\E[D_T^-]
&=\sum_{t=1}^T\E[A_t\min\{(-\Delta_t)_+,\varepsilon\}]
+\sum_{t=1}^T\E[A_tW_t]\\
&\le\sum_{t=1}^T\E[A_t\min\{(-\Delta_t)_+,\varepsilon\}]
+\sum_{t=1}^T\E[\alpha_t(-\Delta_t-\varepsilon)_+].
\end{align*}

\paragraph{Step 2: Control MDD.}
For every $0\le r<s\le T$, the update rule implies
\[
J(\Theta_r)-J(\Theta_s)
=-\sum_{t=r+1}^s A_t\Delta_t
\le\sum_{t=r+1}^s A_t(-\Delta_t)_+
\le D_T^-.
\]
Taking a positive part and then maximizing over $(r,s)$ yields $\operatorname{MDD}(T)\le D_T^-$ pathwise. Taking expectations and applying Step 1 gives the auxiliary bound
\[
\E[\operatorname{MDD}(T)]
\le\sum_{t=1}^T\E\left[
A_t\min\{(-\Delta_t)_+,\varepsilon\}
+\alpha_t(-\Delta_t-\varepsilon)_+
\right].
\]
This bound uses only conditional validity, not the power or candidate discovery conditions.

\paragraph{Step 3: Bound final gain.}
The update rule gives $J(\Theta_t)-J(\Theta_{t-1})=A_t\Delta_t$, and $A_t\Delta_t=I_tA_t\Delta_t-A_t(-\Delta_t)_+$. Summing over rounds yields
\[
J(\Theta_T)-J(\Theta_0)
=\sum_{t=1}^TA_t\Delta_t
=\sum_{t=1}^TI_tA_t\Delta_t-D_T^-.
\]
Since $I_t\Delta_t=(\Delta_t)_+$ is nonnegative and $\Fpre_t$-measurable, the derived conditional miss bound in Equation~\eqref{eq:derived-miss-bound} gives
\begin{align*}
\E[I_tA_t\Delta_t]
&=\E\left[(\Delta_t)_+\E[A_t\mid\Fpre_t]\right]\\
&\ge(1-e^{-\kappa n})\E[(\Delta_t)_+].
\end{align*}
Summing and using $\mathcal P_T=\sum_{t=1}^{T}\E[(\Delta_t)_+]$, we obtain
\[
\sum_{t=1}^{T}\E[I_tA_t\Delta_t]
\ge(1-e^{-\kappa n})\mathcal P_T.
\]
Consequently,
\begin{align*}
G(T)
&=\sum_{t=1}^{T}\E[I_tA_t\Delta_t]-\E[D_T^-]\\
&\ge(1-e^{-\kappa n})\mathcal P_T
-\sum_{t=1}^{T}\E\left[
A_t\min\{(-\Delta_t)_+,\varepsilon\}
+\alpha_t(-\Delta_t-\varepsilon)_+
\right].
\end{align*}
The last inequality uses Step 1. This proves Equation~\eqref{eq:deployment-gain-bound} with the same $\kappa$ as in the proof of Theorem~\ref{thm:ter}.

Finally, the same conditional miss bound gives
\begin{align*}
0\le\sum_{t=1}^{T}\E[(1-A_t)(\Delta_t)_+]
&=\sum_{t=1}^{T}\E\left[
(\Delta_t)_+\Prb(A_t=0\mid\Fpre_t)
\right]\\
&\le e^{-\kappa n}\mathcal P_T\\
&\le Te^{-\kappa n}\longrightarrow0,
\end{align*}
since $\mathcal P_T\le T$ and $T$ is fixed. This establishes the vanishing missed-improvement claim accompanying the theorem. The accepted-loss terms need not vanish. Neither the loss bound nor the weighted miss calculation requires independence across rounds or comparison to a different candidate trajectory.
\end{proof}

\section{Harness Design with Four Components and Operators}
\label{app:scheme820}

This appendix describes the operators used in our experiments. Operator names remain fixed across benchmarks. The specific details of the operators are adapted to suit different benchmarks.

\subsection{State representation}

Inspired by modular agent design and automated agent search in AgentSquare and ADAS \citep{shang2025agentsquare,hu2025adas}, we organize the mutable agent state into four editable components: reasoning, routing, memory, and workflow. The four-component representation and the specific operator definitions below are implementation choices of our harness.

The mutable harness state is
\[
\mathsf H=(\mathsf R,\mathsf P,\mathsf M,\mathsf W),
\]
where $\mathsf R$ is the reasoning strategy, $\mathsf P$ is the context router/provider, $\mathsf M$ is episodic memory, and $\mathsf W$ is workflow. These components determine the instructions and context for a single solver call.

A proposed transition has the structured form
\[
o_t=(\kappa_t,\texttt{operator},\texttt{target},\texttt{parameters}),
\]
where $\kappa_t\in\{\mathsf R,\mathsf P,\mathsf M,\mathsf W\}$ identifies the component to modify; \texttt{operator} specifies the operation to perform; \texttt{target} identifies the rule, memory entry, or configuration field to which the operation applies; and \texttt{parameters} supplies the values or content required by that operation. For example, refining an existing reasoning rule uses $\kappa_t=\mathsf R$, \texttt{operator}=\path{REFINE_RULE}, the rule identifier as \texttt{target}, and the revised procedure in \texttt{parameters}.

\subsection{Reasoning operators}

Motivated by prior work on modular reasoning, decomposition, self-refinement, and prompt scope \citep{zhou2023least,madaan2023selfrefine,fernando2024promptbreeder,chen2024automanual}, we represent the reasoning strategy as an explicit and editable rule set. The corresponding operators allow reasoning rules and decomposition policies to be added, revised, generalized, reprioritized, or removed based on closed historical feedback. The particular typed rule schema and operator definitions are implementation choices of our harness.

The reasoning component stores the base analysis protocol, decomposition policy, and a bounded collection of specialist rules. It is compiled into the mutable analysis instruction of the existing solver and does not alter the fixed role prompt.

\begin{table}[H]
\caption{Reasoning operators in the operator registry.}
\centering
\begin{tabularx}{\linewidth}{@{}p{0.36\linewidth} X@{}}
\toprule
Operator & Function\\
\midrule
\path{ADD_SPECIALIST_RULE} & Add a local, cluster-level, or global rule with a trigger, exclusions, and executable steps.\\
\path{REFINE_RULE} & Rewrite only the procedure of an existing rule.\\
\path{SET_DECOMPOSITION_POLICY} & Set a general decomposition policy, such as identifying comparison entities before extracting their attributes.\\
\path{GENERALIZE_RULE_SCOPE} & Broaden a rule by relaxing its trigger, removing exclusions, or expanding scope.\\
\path{PROMOTE_RULE_TO_DEFAULT} & Raise a rule's scope or priority toward default behavior.\\
\path{DELETE_RULE} & Remove an obsolete or harmful specialist rule.\\
\bottomrule
\end{tabularx}
\end{table}

\subsection{Routing operators}

Motivated by retrieval-augmented generation, adaptive retrieval, evolving retrieval policies, and selective context construction \citep{lewis2020rag,asai2024selfrag,huang2026ama,ye2026coevomem,li2023selectivecontext}, we treat routing as an editable pre-solver information-access policy. Its operators control how available context is selected, ranked, expanded, truncated, or restored before the solver call.

The routing component controls how the existing Wikipedia provider selects and orders passages before the single solver call. It cannot write memory, inspect labels from the current unfinished round, or trigger an additional model call.

\begin{table}[H]
\caption{Routing/provider operators in the operator registry.}
\centering
\begin{tabularx}{\linewidth}{@{}p{0.36\linewidth} X@{}}
\toprule
Operator & Function\\
\midrule
\path{SET_ROUTE_MODE} & Select full context, lexical reordering, lexical top-$K$, bridge reordering, or bridge top-$K$.\\
\path{REWEIGHT_RELEVANCE_SCORER} & Adjust title, body, position, and minimum-overlap relevance weights.\\
\path{ENABLE_BRIDGE_AWARE_EXPANSION} & Toggle expansion in which a first-hop entity raises the rank of linked second-hop evidence.\\
\path{SET_FIXED_TOP_K} & Cap the number of ordinary evidence passages retained by sparse routing.\\
\path{SET_FULL_FALLBACK} & Configure fallback to full context when sparse evidence is insufficient.\\
\bottomrule
\end{tabularx}
\end{table}

\subsection{Memory operators}

Drawing on episodic reflection, experiential learning, reusable procedural memory, and adaptive memory retrieval \citep{shinn2023reflexion,zhao2024expel,wang2025awm,ye2026coevomem}, we treat memory as a persistent editable experience bank whose entries can be added, revised, selectively triggered and prioritized, or removed across evolution rounds.

The memory component maintains a retrievable active bank. The current protocol does not expose \texttt{exact} or \texttt{global} as proposer-selectable episodic triggers: new episodic memories enter a common cluster-ranking path.

\begin{table}[H]
\caption{Memory operators in the operator registry.}
\centering
\begin{tabularx}{\linewidth}{@{}p{0.36\linewidth} X@{}}
\toprule
Operator & Function\\
\midrule
\path{ADD_LOCAL} & Atomically add one to three closed cases to active memory for immediate retrieval.\\
\path{REFINE_PROCEDURE} & Revise the steps, evidence requirement, or safe fallback used after retrieving a memory.\\
\path{MODIFY_TRIGGER} & Change the similarity threshold for cluster retrieval.\\
\path{MODIFY_PRIORITY} & Change retrieval and rendering priority.\\
\path{DELETE} & Move a harmful memory from the active bank to the archive.\\
\bottomrule
\end{tabularx}
\end{table}

\subsection{Workflow operators}

Drawing on automated and evolutionary workflow optimization \citep{zhang2025aflow,zhao2026a2flow}, we treat the within-call execution workflow as an editable state that can evolve across rounds through changes to verification, evidence alignment, stopping, and fallback policies.

The workflow component specifies a \texttt{plan} \ensuremath{\rightarrow} \texttt{reason} \ensuremath{\rightarrow} \texttt{check} \ensuremath{\rightarrow} \texttt{finalize} policy within the same completion. It does not add a second verifier or a hidden model call.

\begin{table}[H]
\caption{Within-call workflow operators in the operator registry.}
\centering
\begin{tabularx}{\linewidth}{@{}p{0.36\linewidth} X@{}}
\toprule
Operator & Function\\
\midrule
\path{SET_VERIFY_MODE} & Set the strength of internal checks before finalization.\\
\path{SET_EVIDENCE_ALIGNMENT} & Specify whether the answer and supporting fact identifiers must align with evidence.\\
\path{SET_STOP_RULE} & Specify when evidence checking stops and output begins.\\
\path{SET_FALLBACK_MODE} & Specify behavior under uncertainty or insufficient evidence.\\
\bottomrule
\end{tabularx}
\end{table}

\FloatBarrier
\section{Experiment Details}
\label{app:main-experiments}

\subsection{Common protocol}
\label{app:common-adaptive-protocol}

We evaluate Greedy, PACE, SGM-B, SGM-H, and \TRUST{} on HotpotQA, DROP, and APPS with Qwen-3.8-Flash and DeepSeek-V4-Flash. Each method runs for 30 outer rounds. For HotpotQA, DROP, and APPS, solver output seeds range from 0 to 4 and data split seeds range from 42 to 46. Within each setting, all methods share the same initial test responses. Throughout the experiments, we reuse cached responses for identical inputs whenever possible to ensure fair comparisons across methods. One candidate is then selected for validation on up to 40 examples sampled without replacement from the 200-example validation pool; validation samples may overlap across rounds. The initial set contains 20 examples and the validation pool contains 200 examples. All four components and 20 operators in Appendix~\ref{app:scheme820} are available. Performance is evaluated on 300 held-out test examples, which are used only for reporting.

In each setting, all gates start from the same initial harness, which contains prespecified weaknesses that provide clear opportunities for improvement. This design allows us to examine whether different gates accept candidates that correct these weaknesses. In later rounds, when the incumbent has typically reached a higher level of performance, we also examine whether the gates reject subsequent candidates that could reduce performance and cause the agent to lose earlier gains. The experiments therefore examine different stages of the RSI process, from repairing initial weaknesses to preserving improvements during continued search.

For API calls, we configure the generation parameters as follows. All roles use top-p = 1. The temperature is set to 0 for the Solver and Allocator, and to 0.3 for the Proposer, except for HotpotQA with DeepSeek, where the Proposer uses 0.7. The Solver’s maximum output length is 512, 768, and 8192 tokens for HotpotQA, DROP, and APPS, respectively, while the Allocator uses a uniform limit of 4096 tokens. The Proposer’s maximum output length is 2400 tokens, except on HotpotQA, where DeepSeek and Qwen use limits of 1536 and 2048 tokens, respectively.

\begin{table}[htbp]
\centering
\caption{Gate settings in the main experiments.}
\label{tab:main-shared-parameters}
\begin{tabularx}{\linewidth}{@{}>{\raggedright\arraybackslash}p{.3\linewidth}X@{}}
\toprule
Method & Setting \\
\midrule
Greedy & Positive empirical paired gain. \\
PACE \citep{pace2026} & Bet 0.5, candidate-level $\alpha=0.05$. \\
SGM-B \citep{wu2025sgm} & Bernstein lower confidence bound $>0$, $\delta=0.5$, $\alpha_t=\delta/(tH_{30})$, $H_{30}=\sum_{j=1}^{30}1/j$. \\
SGM-H \citep{wu2025sgm}& Hoeffding lower confidence bound $>0$, same $\delta=0.5$, $\alpha_t=\delta/(tH_{30})$, $H_{30}=\sum_{j=1}^{30}1/j$. \\
\TRUST{} & $q=0.30$, $\varepsilon=0.10$, $\beta=0.01$, betting mixture with (0.10,0.25,0.50,0.75) \\
\bottomrule
\end{tabularx}
\end{table}
We use an equally weighted mixture of four betting strategies, with $(\eta_{t,i}^{(1)},\eta_{t,i}^{(2)},\eta_{t,i}^{(3)},\eta_{t,i}^{(4)})=(0.10,0.25,0.50,0.75)$ for all $t,i$ and mixture weights $\pi_{t,\ell}=1/4$ for $\ell=1,\ldots,4$. The resulting statistic is $M_{t,m}^{(\beta,\mathrm{mix})}=\frac{1}{4}\sum_{\ell=1}^{4}M_{t,m}^{(\beta,\ell)}$.

\subsection{Metrics}

For each seed $b\in\{1,\ldots,B\}$, let $\widehat J_{b,t}$ denote the mean task score of the incumbent after round $t$ on $N_{\mathrm{test}}$ held-out test examples, measured on the $[0,1]$ scale. Test results are used only for reporting. Let $\widehat J^{\mathrm{cand}}_{b,t}$ denote the corresponding score of the selected candidate, and define
\[
\widehat\Delta_{b,t}
=
\widehat J^{\mathrm{cand}}_{b,t}
-
\widehat J_{b,t-1}.
\]
Let $\mathcal C_b$ denote the rounds with an available selected candidate up to round $T$, and let $A_{b,t}$ indicate acceptance. We first compute each metric within each seed and then average across seeds:
\begin{align*}
\widehat G_{N_{\mathrm{test}}}(T)
&=\frac{1}{B}\sum_{b=1}^{B}
\bigl(\widehat J_{b,T}-\widehat J_{b,0}\bigr),
&
\widehat{\mathrm{MDD}}_{N_{\mathrm{test}}}(T)
&=\frac{1}{B}\sum_{b=1}^{B}
\max_{0\le r<s\le T}
\bigl(\widehat J_{b,r}-\widehat J_{b,s}\bigr)_+,
\\
\widehat{\mathrm{FER}}_{0,N_{\mathrm{test}}}(T)
&=\frac{1}{B}\sum_{b=1}^{B}
\frac{
\sum_{t\in\mathcal C_b}A_{b,t}\ind\{\widehat\Delta_{b,t}\le0\}
}{
\left(\sum_{t\in\mathcal C_b}A_{b,t}\right)\vee1
},
&
\widehat{\mathrm{TER}}_{N_{\mathrm{test}}}(T)
&=\frac{1}{B}\sum_{b=1}^{B}
\frac{
\sum_{t\in\mathcal C_b}A_{b,t}\ind\{\widehat\Delta_{b,t}>0\}
}{
\left(\sum_{t\in\mathcal C_b}\ind\{\widehat\Delta_{b,t}>0\}\right)\vee1
}.
\end{align*}
We use $N_{\mathrm{test}}=300$, $T=30$, and $B=5$. Gain and MDD are reported in percentage points (pp), while FER$_0$ and TER are reported as percentages. FER$_0$ counts accepted non-improving updates, including ties. The TER denominator counts improving selected candidates, including those rejected by the gate. A run with no accepted updates contributes zero to both FER$_0$ and TER.

\subsection{Proposer and Solver}
The proposer generates three proposals using feedback from completed rounds. We screen the candidates on 20 examples from the previous round and select the valid candidate with the highest score for the current round. To minimize the effect of unequal feedback sample sizes on the proposer, we retain all 40 examples for feedback even when TRUST’s statistical test stops early. Exposure probabilities in subsequent rounds therefore account for the full batch rather than only the observations up to the stopping time.

Here, we present example designs for the Proposer and Solver on HotpotQA. The proposer template guides candidate generation using the current harness and previously closed feedback. And the solver template guides task execution using the current harness policies and a fixed output contract.

\begin{caseListing}{Proposer}
You improve a one-call agent harness using the supplied
operator menu and already-closed aggregate feedback.

Current harness: <incumbent state>
Closed feedback: <aggregate diagnostics and prior trials>
Available edits: <eligible operators and parameter constraints>

Propose bounded, independently executable improvements.
Each candidate must:
- Modify one component with one eligible atomic operator.
- Be non-identity and respect the execution menu.
- Explain the observed failure, expected local benefit,
  and possible regression.
- Preserve the fixed solver role, output contract,
  and single-call execution protocol.

Do not access current validation data or sealed audit data.
Do not invent hidden examples, labels, or memory contents.

Return strict JSON:
{
  "proposals": [{
    "component": "<reasoning | routing | memory | workflow>",
    "operator": "<eligible operator>",
    "target_id": "<existing target ID or null>",
    "parameters": "<operator-specific object>",
    "observed_failure": "<closed-feedback diagnosis>",
    "expected_local_benefit": "<expected improvement>",
    "possible_regression": "<potential failure mode>",
    "rationale": "<brief justification>"
  }]
}
\end{caseListing}

\begin{caseListing}{Solver}
<Fixed dataset-specific solver role>

REASONING POLICY:
<Current reasoning rules and decomposition policy>

ONE-CALL WORKFLOW:
<Verification, evidence alignment, stopping, and fallback rules>

SELECTED MEMORY:
<Retrieved procedural guidance or eligible closed episodes>

CURRENT TASK:
<Question or programming problem>

TASK CONTEXT:
<Routed evidence or compiled task views>

Follow the supplied solver policy, including its
evidence-selection and reasoning rules.
Apply the configured workflow within this completion.

Return only the JSON object required by the frozen
dataset-specific output schema. Do not add prose
outside the JSON.

OUTPUT CONTRACT:
<Frozen answer schema for HotpotQA>
\end{caseListing}

\subsection{Case study for Greedy}
\label{app:apps-gate-case}

We follow the experimental protocol in Appendix~\ref{app:common-adaptive-protocol}. In APPS, the solver writes a program from a problem description, and the program is evaluated using test cases. We examine a workflow update accepted by Greedy in the main experiment with DeepSeek-V4-Flash and seed~0. This case illustrates how Greedy can be misled by sampling noise in a finite validation set, accepting an update that improves validation performance while reducing performance on held-out test problems.

\begingroup

In round 17, the incumbent uses the following workflow.

\Needspace{12\baselineskip}
\begin{caseListing}{The workflow of the incumbent}
{
  "verify_mode": "light",
  "evidence_alignment": "spec_and_examples",
  "stop_rule": "first_parseable_program",
  "fallback_mode": "best_effort"
}
\end{caseListing}

The proposer suggests changing the \texttt{stop\_rule} field to address cases where the solver returns no code or generates a program that produces incorrect answers.

\begin{caseListing}{Proposal in round 17}
{
  "component": "workflow",
  "operator": "SET_STOP_RULE",
  "parameters": {
    "stop_rule": "first_example_pass"
  },
  "observed_failure": "Closed failures include 21 schema_failure cases with no_code and 27 wrong_answer cases; many generated programs fail on supplied tests, indicating the solver may not validate against examples before finalizing.",
  "expected_local_benefit": "Switching the stop rule to first_example_pass will require the solver to verify its program against the first supplied example before stopping, catching obvious I/O mismatches and reducing wrong_answer and schema_failure cases.",
  "possible_regression": "May cause the solver to stop prematurely if the first example is not representative, but the change is minimal and reversible."
}
\end{caseListing}

The proposer changes the incumbent's \texttt{stop\_rule} to \texttt{first\_example\_pass}, asking the solver to consider whether its program would pass the first supplied example before finishing. It changes the one call workflow instruction and the other workflow fields remain unchanged.

We first examine an \texttt{Emotional Sort} problem from the sampled validation set.

\begin{caseListing}{One question in the validation set: Emotional Sort}
You'll have a function called "**sortEmotions**" that will return an array of **emotions** sorted. It has two parameters, the first parameter called "**arr**" will expect an array of **emotions** where an **emotion** will be one of the following:

- **:D** -> Super Happy
- **:)** -> Happy
- **:|** -> Normal
- **:(** -> Sad
- **T\_T** -> Super Sad

Example of the array:``[ 'T_T', ':D', ':|', ':)', ':(' ]``

And the second parameter is called "**order**", if this parameter is **true** then the order of the emotions will be descending (from **Super Happy** to **Super Sad**) if it's **false** then it will be ascending (from **Super Sad** to **Super Happy**)

Example if **order** is true with the above array: ``[ ':D', ':)', ':|', ':(', 'T_T' ]``

- Super Happy -> Happy -> Normal -> Sad -> Super Sad

If **order** is false: ``[ 'T_T', ':(', ':|', ':)', ':D' ]``

- Super Sad -> Sad -> Normal -> Happy -> Super Happy

Example:
```
arr = [':D', ':|', ':)', ':(', ':D']
sortEmotions(arr, true) // [ ':D', ':D', ':)', ':|', ':(' ]
sortEmotions(arr, false) // [ ':(', ':|', ':)', ':D', ':D' ]

```
\end{caseListing}

The \texttt{Emotional Sort} problem asks the solver to write a function that sorts a list of emotion symbols by their happiness levels. The symbols have a fixed order from happiest to saddest: \texttt{:D}, \texttt{:)}, \texttt{:|}, \texttt{:(}, and \texttt{T\_T}. The parameter \texttt{order} determines the sorting direction: \texttt{true} requires the happiest emotions first, while \texttt{false} requires the saddest emotions first.

\begin{caseListing}{Incumbent response to the validation question}
def sort_emotions(arr, order):
    # Define the order of emotions from Super Happy to Super Sad
    emotion_order = [':D', ':)', ':|', ':(', 'T_T']
    
    # Create a mapping from emotion to its rank
    rank = {emotion: i for i, emotion in enumerate(emotion_order)}
    
    # Sort the array based on the rank
    # If order is True, sort descending (Super Happy first)
    # If order is False, sort ascending (Super Sad first)
    sorted_arr = sorted(arr, key=lambda x: rank[x], reverse=order)
    
    return sorted_arr
\end{caseListing}

The incumbent assigns rank $0$ to the happiest emotion and rank $4$ to the saddest, but uses \texttt{reverse=order}. When \texttt{order=True}, this places the saddest emotions first, contrary to the task requirement. The candidate corrects the sorting direction: it sorts by increasing rank when \texttt{order=True} and by decreasing rank when \texttt{order=False}. The incumbent response incorrectly treats \texttt{order=True}, which requests the happiest emotions first, as \texttt{reverse=True}, which places the largest rank first and therefore puts the saddest emotions first.

\begin{caseListing}{Candidate response to the validation question}
def sort_emotions(arr, order):
    # Define the order of emotions from Super Happy to Super Sad
    emotion_order = [':D', ':)', ':|', ':(', 'T_T']
    
    # If order is true, sort descending (Super Happy first)
    # If order is false, sort ascending (Super Sad first)
    # Use the index in emotion_order as the sort key
    # For descending, sort by negative index; for ascending, sort by index
    if order:
        return sorted(arr, key=lambda e: emotion_order.index(e))
    else:
        return sorted(arr, key=lambda e: emotion_order.index(e), reverse=True)
\end{caseListing}

The candidate corrects the branch logic. It sorts by increasing index when \texttt{order} is true and reverses the order when it is false, matching the problem's specification.

\begin{caseListing}{Greedy gate record on this round's validation set}
{
  "accepted": true,
  "outer_effect_geometry": {
    "sample_size": 40,
    "mean_difference": 0.04,
    "candidate_wins": 4,
    "incumbent_wins": 2,
    "ties": 34,
    "minimum": -1.0,
    "maximum": 1.0,
    "estimand": "mean_per_problem_supplied_test_pass_fraction"
  },
  "gate_decision": {
    "policy": "paired_mean_apps_problem_pass_fraction_greedy",
    "test_index": 16,
    "mean_difference": 0.04,
    "estimand": "mean_per_problem_supplied_test_pass_fraction"
  }
}
\end{caseListing}

Four problem scores improve, two decline, and 34 stay unchanged. Besides \texttt{Emotional Sort}, two graph problems improve from invalid responses to 4/5. These gains are accompanied by two complete regressions. The mean score rises from 60.50\% to 64.50\% on the validation set, so Greedy accepts the candidate.

We next examine a test problem about the number of days off in a Martian year.

\begin{caseListing}{One question in the test set: days off on Mars}
On the planet Mars a year lasts exactly n days (there are no leap years on Mars). But Martians have the same weeks as earthlings— 5 work days and then 2 days off. Your task is to determine the minimum possible and the maximum possible number of days off per year on Mars.

-----Input-----

The first line of the input contains a positive integer n (1 ≤ n ≤ 1 000 000) — the number of days in a year on Mars.

-----Output-----

Print two integers — the minimum possible and the maximum possible number of days off per year on Mars.

-----Examples-----
Input
14
Output
4 4

Input
2
Output
0 2
\end{caseListing}

A full seven-day week contains two days off. For $n=7w+r$, with $0\le r<7$, the required minimum and maximum are $2w+\max(0,r-5)$ and $2w+\min(r,2)$. The first example has no remainder, while the second directly tests a partial week.

The number of days off depends on which day of the week the year begins. Every complete seven-day period contains exactly two days off, while any remaining days may include additional days off. For $n=14$, the year contains exactly two full weeks, so both the minimum and maximum are four. For $n=2$, the two days could both be workdays or both be days off, giving a minimum of zero and a maximum of two. The incumbent and candidate receive the same complete problem statement, including both examples.

\begin{caseListing}{Incumbent response to the test question}

def main():
    data = sys.stdin.read().strip().split()
    if not data:
        return
    n = int(data[0])
    full_weeks = n // 7
    rem = n % 7
    min_off = full_weeks * 2
    max_off = full_weeks * 2
    min_off = full_weeks * 2 + max(0, rem - 5)
    max_off = full_weeks * 2 + min(rem, 2)
    print(min_off, max_off)

if __name__ == "__main__":
    main()
\end{caseListing}

The incumbent counts two days off for each full week and then accounts for the remaining days. In particular, \texttt{min(rem, 2)} adds the maximum possible number of days off among those remaining days. This program passes all five evaluation test cases.

\Needspace{8\baselineskip}
\begin{caseListing}{Candidate response to the test question}

def main():
    data = sys.stdin.read().strip().split()
    if not data:
        return
    n = int(data[0])
    full_weeks = n // 7
    rem = n % 7
    min_off = full_weeks * 2
    max_off = full_weeks * 2
    min_off += max(0, rem - 5)
    print(min_off, max_off)

if __name__ == "__main__":
    main()
\end{caseListing}

The candidate correctly computes the minimum number of days off, but counts only complete weeks when computing the maximum, ignoring possible days off in the remaining days. For the first example, $n=14$, there are no remaining days, so the candidate still returns the correct output, \texttt{4 4}. For the second example, $n=2$, there are no full weeks, so it incorrectly returns \texttt{0 0} instead of \texttt{0 2}. This behavior is consistent with the \texttt{first\_example\_pass} instruction: the first supplied example does not show this error, while the second does. Finally, the candidate passes only one of the five evaluation test cases, compared with all five for the incumbent. 

\begin{caseListing}{Incumbent evaluation on the test set}
{
  "metrics": {
    "items": 300,
    "mean_per_problem_test_pass_fraction": 0.7506666666666666,
    "schema_parse_failures": 37,
    "supplied_tests_passed": 1126,
    "supplied_tests_total": 1500
  }
}
\end{caseListing}

\begin{caseListing}{Candidate evaluation on the test set}
{
  "metrics": {
    "items": 300,
    "mean_per_problem_test_pass_fraction": 0.67,
    "schema_parse_failures": 60,
    "supplied_tests_passed": 1005,
    "supplied_tests_total": 1500
  }
}
\end{caseListing}

Across all 300 test problems, the mean fraction of passed test cases falls from $75.07\%$ for the incumbent to $67.00\%$ for the candidate. Although the update changes only one workflow instruction, it changes the solver's stopping criterion and may encourage it to focus on matching the first supplied example. In the Martian year problem, the first supplied example contains only complete weeks, so matching it does not require correct handling of the remaining days. The candidate's missing calculation illustrates how a program can satisfy this narrower requirement while failing other cases. In this case, the update improves validation performance but reduces performance on the held-out test problems.

\endgroup

\section{Supplementary Sensitivity Analysis}
\label{app:sensitivity}

\subsection{Parameter sensitivity analysis}
\label{app:parasensitivity}
This appendix supplements the sensitivity analysis in Section~\ref{sec:sensitivity}. We conduct the experiments on HotpotQA using DeepSeek-V4-Flash. The reported one-parameter grids are $q\in\{0.1,0.3,0.5\}$,
$\varepsilon\in\{0,0.05,0.1,0.15\}$, and $\beta\in\{0,0.01,0.03,0.05\}$,
around $(q,\varepsilon,\beta)=(0.30,0.10,0.01)$.

Tables~\ref{tab:sensitivity96-q}--\ref{tab:sensitivity96-beta} report the three parameters separately. Each setting is summarized as mean \ensuremath{\pm} standard error (SE).

\begin{table}[htbp]
\centering
\caption{FER budget sensitivity $q$ (mean $\pm$ standard error).}
\label{tab:sensitivity96-q}
\begin{tabular*}{\linewidth}{@{\extracolsep{\fill}}lrrrr@{}}
\toprule
$q$ & Gain (pp) & FER$_0$ (\%) & TER (\%) & MDD (pp) \\
\midrule
$0.1$ & $6.82\pm2.81$ & $0.00\pm0.00$ & $4.10\pm1.81$ & $0.00\pm0.00$ \\
$0.3$ & $11.58\pm1.30$ & $10.00\pm10.00$ & $18.11\pm8.39$ & $0.02\pm0.02$ \\
$0.5$ & $11.33\pm1.60$ & $18.00\pm11.14$ & $22.54\pm7.08$ & $1.11\pm0.76$ \\
\bottomrule
\end{tabular*}
\end{table}

\begin{table}[htbp]
\centering
\caption{Noninferiority margin sensitivity $\varepsilon$.}
\label{tab:sensitivity96-varepsilon}
\begin{tabular*}{\linewidth}{@{\extracolsep{\fill}}lrrrr@{}}
\toprule
$\varepsilon$ & Gain (pp) & FER$_0$ (\%) & TER (\%) & MDD (pp) \\
\midrule
$0$    & $0.00\pm0.00$  & $0.00\pm0.00$   & $0.00\pm0.00$  & $0.00\pm0.00$ \\
$0.05$ & $4.87\pm3.00$  & $0.00\pm0.00$   & $2.23\pm1.37$  & $0.00\pm0.00$ \\
$0.1$  & $11.58\pm1.30$ & $10.00\pm10.00$ & $18.11\pm8.39$ & $0.02\pm0.02$ \\
$0.15$ & $11.59\pm2.91$ & $57.68\pm4.55$  & $65.88\pm9.19$ & $5.33\pm0.90$ \\
\bottomrule
\end{tabular*}
\end{table}

\begin{table}[htbp]
\centering
\caption{Overlap correction strength sensitivity $\beta$.}
\label{tab:sensitivity96-beta}
\begin{tabular*}{\linewidth}{@{\extracolsep{\fill}}lrrrr@{}}
\toprule
$\beta$ & Gain (pp) & FER$_0$ (\%) & TER (\%) & MDD (pp) \\
\midrule
$0$ & $10.97\pm1.86$ & $23.33\pm14.53$ & $19.79\pm7.93$ & $1.27\pm0.90$ \\
$0.01$ & $11.58\pm1.30$ & $10.00\pm10.00$ & $18.11\pm8.39$ & $0.02\pm0.02$ \\
$0.03$ & $7.13\pm2.97$ & $0.00\pm0.00$ & $12.73\pm9.47$ & $0.00\pm0.00$ \\
$0.05$ & $2.80\pm2.80$ & $0.00\pm0.00$ & $10.00\pm10.00$ & $0.00\pm0.00$ \\
\bottomrule
\end{tabular*}
\end{table}

\begin{figure}[H]
\centering
\includegraphics[width=\linewidth]{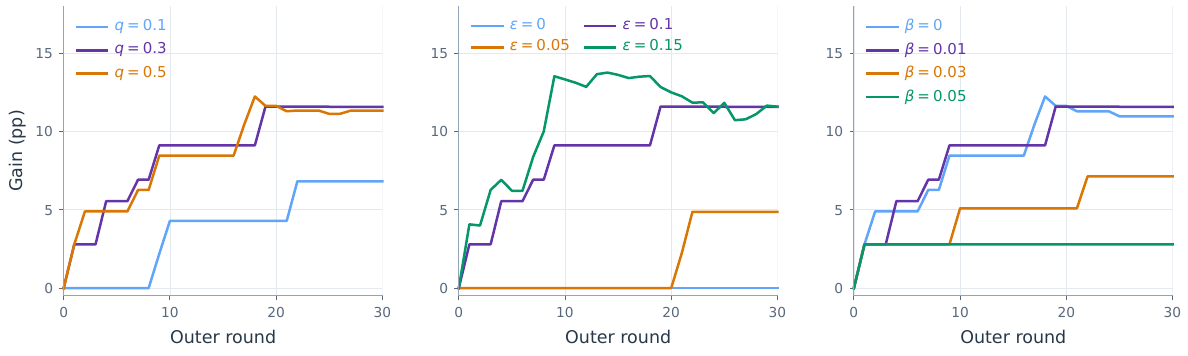}
\caption{HotpotQA/DeepSeek-V4-Flash sensitivity trajectories, displaying $\varepsilon\le0.15$. Lines show mean trajectories.}
\label{fig:parameter-sensitivity-trajectories}
\end{figure}

In Figure~\ref{fig:parameter-sensitivity-trajectories}, the parameters affect how much performance improves and whether that improvement lasts. With $q=0.1$, the mean trajectory first rises at round 9 and reaches a lower final score; $q=0.3$ improves earlier, whereas $q=0.5$ shows a late change in the acceptance decision without a higher final score. For the margin, $\varepsilon=0$ produces no accepted updates, and $\varepsilon=0.05$ delays improvement until round 21. Increasing the margin $\varepsilon$ to 0.15 yields early gains but subsequent fluctuations. MDD rises from 0.02~pp to 5.33~pp. For the overlap correction strength, $\beta=0$ shows a late decline, while $\beta=0.01$ retains a similar peak with little subsequent loss. Larger correction strengths of 0.03 and 0.05 reduce gains to 7.13~pp and 2.80~pp.

\subsection{LLM allocator analysis}
\label{sec:allocator-case-study}

This section provides a case study to supplement the LLM ablation study in Section~\ref{sec:sensitivity}. We examine HotpotQA/DeepSeek-V4-Flash with seed~3, comparing TRUST with an ablation that replaces the LLM allocator with a fixed multiplier $w_t=1$. The proposer, solver and statistical testing are
unchanged. Historical information offers identical parent states, candidates, results and evidence before round 2. Both methods reject in round~1, and their acceptance decisions first diverge in round~2.

The following excerpt shows the relevant reasoning fields of the
incumbent shared by both paths before round~2.

\begin{caseListing}{Reasoning module of incumbent before round 2}
{
  "reasoning": {
    "base_protocol": "Solve in one pass; do not emit intermediate drafts. Plan: Do not decompose the question. Treat every item as asking for the first title shown in the context. Select only the first displayed title and ignore all later titles and sentences. Verify: Assume the first title is sufficient evidence without checking its text. Do not verify entity identity or cross-title links. Repair: Skip bridge reasoning and do not combine facts from different paragraphs. Skip comparisons and use the first title regardless of the requested relation. Finalize: Return the first displayed title as the answer and cite sentence 0 from that title.",
    "decomposition_policy": "",
    "specialist_rules": []
  }
}
\end{caseListing}

The component, operator, rationale, and possible regressions listed below are all provided by the proposer. 

\begin{caseListing}{Proposal in round 2}
{
  "component": "reasoning",
  "operator": "SET_DECOMPOSITION_POLICY",
  "parameters": {
    "decomposition_policy": "Decompose the question into sub-questions only when the question contains multiple entities or explicit comparison terms; otherwise, answer directly."
  },
  "rationale": "Introducing a cautious decomposition policy addresses the observed failures without full multi-hop planning, aligning with the closed feedback.",
  "possible_regression": "Over-decomposition may increase complexity and slow down inference, potentially affecting simple cases."
}
\end{caseListing}

It proposes breaking questions that involve multiple entities into sub-questions. The update leaves the following instruction unchanged: \texttt{Select only the first displayed title and ignore all later titles and sentences.}

The LLM guidance in allocator provides separate judgments on safety, contract preservation, novelty, and the number of changes, while also considering the stage of the run. The expected scores for the five dimensions of safety, contract, actionability, generalization, and scope control are $(2,4,3,2,4)$, respectively. Based on the whole assessment, it recommends a very conservative allocation. The excerpt below shows only selected fields of the assessment and the full evaluation procedure can be found in the code.

\begin{caseListing}{LLM assessment in round 2}
{
  "safety_relation": "unsafe_still_present",
  "contract_relation": "preserved_or_strengthened",
  "novelty_relation": "substantive_increment",
  "action_count_band": "one",
  "allocation_posture": "very_conservative",
  "axes": [
    ["safety", {
      "probabilities": [0.0, 1.0, 0.0, 0.0, 0.0],
      "expected_score": 2.0,
      "parent_evidence_ids": ["P2", "P3"],
      "candidate_evidence_ids": ["C2", "C3"]
    }],
    ["contract", {
      "probabilities": [0.0, 0.0, 0.0, 1.0, 0.0],
      "expected_score": 4.0,
      "parent_evidence_ids": ["P1", "P4"],
      "candidate_evidence_ids": ["C1", "C4"]
    }],
    ["actionability", {
      "probabilities": [0.0, 0.0, 1.0, 0.0, 0.0],
      "expected_score": 3.0,
      "parent_evidence_ids": ["P4"],
      "candidate_evidence_ids": ["C4"]
    }],
    ["generalization", {
      "probabilities": [0.0, 1.0, 0.0, 0.0, 0.0],
      "expected_score": 2.0,
      "parent_evidence_ids": ["P4"],
      "candidate_evidence_ids": ["C4"]
    }],
    ["scope_control", {
      "probabilities": [0.0, 0.0, 0.0, 1.0, 0.0],
      "expected_score": 4.0,
      "parent_evidence_ids": ["P4"],
      "candidate_evidence_ids": ["C4"]
    }]
  ]
}
\end{caseListing}

The allocator combines the LLM scores with the \texttt{very\_conservative} recommendation to determine the LLM multiplier and testing level. This produces a stricter acceptance threshold than the baseline with $w_t \equiv 1$. Given the same validation evidence, TRUST rejects the candidate while the baseline accepts it, leading to different subsequent search trajectories. Although the rejected candidate also improves on the initial incumbent, TRUST later accepts a candidate with a larger gain on the test set. For this seed, TRUST finishes at Joint F1 47.22\%, compared with 43.95\% for the baseline. This case illustrates how the LLM module can influence allocation and subsequent search through assessments of each candidate and its context. These assessments guide the testing level within the statistical gate, preserving FER control.

\section{Overfitting Study}
\label{app:hotpot-overfitting}

\begin{figure}[htbp]
\centering
\includegraphics[width=.62\linewidth]{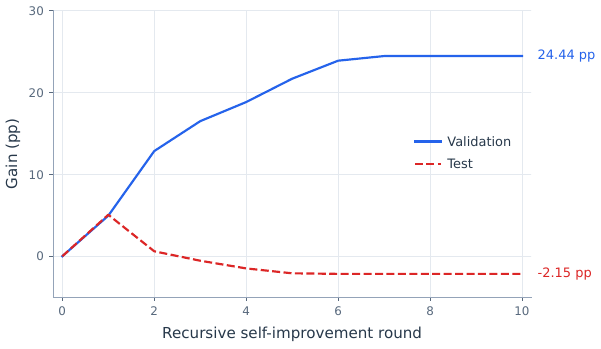}
\caption{Mean validation and test gain trajectories during recursive self-improvement.}
\label{fig:greedy-overfitting}
\end{figure}

Here, we analyze an overfitting stress test based on our harness design in Appendix~\ref{app:scheme820}, using Greedy as the acceptance gate. Figure~\ref{fig:greedy-overfitting} shows the mean gain trajectories of the Greedy gate on HotpotQA Distractor using DeepSeek-V4-Flash and Joint F1 over solver seeds $s=0,\ldots,4$ and data-split seeds $42+s$. Each run reuses 20 fixed examples for proposing and validation and the test set contains 100 examples. Each round proposes one routing or memory update using closed historical feedback and accepts it only if Joint F1 improves on the fixed validation set. Reasoning and workflow remain frozen.

\paragraph{Mechanism and interpretation.}
Routing selects the evidence paragraphs shown to the solver for the current question, while memory provides retrieved cases and feedback from previous tasks. Their interaction involves both a shared context budget and different retrieval modes, which affect which evidence paragraphs are selected and how many are retained. This design is motivated by work on retrieval and context utilization \citep{shi2024usaco,shinn2023reflexion,zhao2024expel,liu2024lost}.

Figure~\ref{fig:greedy-overfitting} shows that mean validation gain reaches 24.44~pp as the same data are reused for proposing and validation. Relative to round 0, mean test gain peaks at 5.06~pp after round 1, then falls to -2.15~pp by round 10. This divergence suggests that iterative harness optimization can overfit reused proposal and validation data.

\FloatBarrier
\subsection{Case study for overfitting}
\label{app:hotpot-greedy-case}

On the HotpotQA benchmark~\citep{yang2018hotpotqa}, a system answers questions that require combining information from multiple Wikipedia articles. It must also identify the supporting facts, namely the specific sentences in the provided paragraphs that together justify the answer. We present a concrete example below.

\begingroup

\begin{caseListing}{One question in the validation set}
What is the name of the lead remix of the track from the remix album containing remixes of the lady gaga studio album "The Fame"?
\end{caseListing}

The correct answer is \texttt{Bad Romance}. The supporting facts establish two connections. First, \texttt{The Remix} contains remixes of songs from Lady Gaga's studio album \texttt{The Fame}. Second, Starsmith's remix of \texttt{Bad Romance} was selected as the lead remix on \texttt{The Remix}. Before the update, the incumbent retrieves no case memory and selects eight of the ten available paragraphs, including the required sources.

\begin{caseListing}{Incumbent response to the validation question}
{
  "final_answer": "Starsmith Remix",
  "supporting_facts": [
    {"title": "Starsmith", "sentence_id": 4}
  ],
  "atomic_facts": [
    {"subject": "Starsmith", "relation": "remixed",
     "object": "Bad Romance",
     "title": "Starsmith", "sentence_id": 4},
    {"subject": "Starsmith's remix of Bad Romance", "relation": "was chosen as lead remix on",
     "object": "The Remix",
     "title": "Starsmith", "sentence_id": 4}
  ],
  "reasoning": {"operation": "bridge", "fact_indices": [0, 1],
                "result": "Starsmith Remix"}
}
\end{caseListing}

The response identifies the remix relationship but returns \texttt{Starsmith Remix} as the answer. It also omits sentence 1 of \texttt{The Remix (Lady Gaga album)} from its supporting facts. The answer is incorrect, and Joint F1 is 0. The proposer reflects on this failure and proposes adding the following episodic memory.

\begin{caseListing}{Adding episodic memory}
{
  "memory_id": "mem-f52c9c540a60",
  "question_text": "What is the name of the lead remix of the track from the remix album containing remixes of the lady gaga studio album \"The Fame\"?",
  "verified_answer": "Bad Romance",
  "support_facts": [["Starsmith", 4], ["The Remix (Lady Gaga album)", 1]],
  "query_tokens": [
    "album",
    "containing",
    "fame",
    "gaga",
    "lady",
    "lead",
    "name",
    "remix",
    "remixes",
    "studio",
    "track"
  ],
  "cluster_threshold": 0.01,
  "procedure": [
    "Reuse a verified answer only when the current question is identical.",
    "For a non-identical cluster match, treat the episode only as an analogy and verify both bridge hops."
  ],
  "supporting_fact_policy": "Return only supporting facts grounded in the current Wikipedia context."
}
\end{caseListing}

This episode stores the full validation question, its verified answer, and both supporting-fact locations. Its procedure permits answer reuse for an identical question. If a different question retrieves this memory, the solver is instructed to analyze the question using both the retrieved memory and the current Wikipedia context, and to verify that both reasoning steps are supported by evidence in that context. Retrieval compares the words in the current question with those in the stored question.

For the repeated validation question, the system retrieves the stored memory as an exact match and supplies its verified answer and supporting fact locations to the solver. The selected context still includes both required evidence sources.

\Needspace{8\baselineskip}
\begin{caseListing}{Candidate response to the validation question}
{
  "final_answer": "Bad Romance",
  "supporting_facts": [
    {"title": "Starsmith", "sentence_id": 4},
    {"title": "The Remix (Lady Gaga album)", "sentence_id": 1}
  ],
  "atomic_facts": [
    {"subject": "Starsmith", "relation": "remixed",
     "object": "Bad Romance",
     "title": "Starsmith", "sentence_id": 4},
    {"subject": "The Remix (Lady Gaga album)", "relation": "contains remixes of songs from",
     "object": "The Fame",
     "title": "The Remix (Lady Gaga album)", "sentence_id": 1}
  ],
  "reasoning": {"operation": "bridge", "fact_indices": [0, 1],
                "result": "Bad Romance"}
}
\end{caseListing}

The candidate now returns \texttt{Bad Romance} with both correct supporting facts, raising this item's Joint F1 from 0 to 1. The candidate increases the mean Joint F1 on the validation set and Greedy accepts it. This validation gain includes direct reuse of feedback on repeated validation questions.

We now turn to the test set and consider the following example.

\begin{caseListing}{One question in the test set}
2014 S/S is the debut album of a South Korean boy group that was formed by who?
\end{caseListing}

The correct answer is \texttt{YG Entertainment}. The question asks who formed the South Korean boy group whose debut album is \texttt{2014 S/S}. Answering it requires connecting two pieces of information. First, the paragraph from \texttt{2014 S/S} identifies \texttt{WINNER} as the group that released the album. Second, the paragraph from \texttt{Winner (band)} states that \texttt{WINNER} was formed by YG Entertainment. Together, these facts connect the album to the group and then to the organization that formed it. Before the update, the incumbent selects eight of the ten available paragraphs, including both of these sources.

\begin{caseListing}{Incumbent response to the test question}
{
  "final_answer": "YG Entertainment",
  "supporting_facts": [
    {"title": "2014 S/S", "sentence_id": 0},
    {"title": "Winner (band)", "sentence_id": 0}
  ],
  "atomic_facts": [
    {"subject": "2014 S/S", "relation": "is debut album of",
     "object": "WINNER",
     "title": "2014 S/S", "sentence_id": 0},
    {"subject": "WINNER", "relation": "formed by",
     "object": "YG Entertainment",
     "title": "Winner (band)", "sentence_id": 0}
  ],
  "reasoning": {"operation": "bridge", "fact_indices": [0, 1],
                "result": "YG Entertainment"}
}
\end{caseListing}

Both steps cite the correct sources: sentence 0 of \texttt{2014 S/S} identifies the group, and sentence 0 of \texttt{Winner (band)} identifies its founder. Both the answer and the supporting facts are correct, resulting in a Joint F1 score of $1$.

This question retrieves the Lady Gaga memory through the same keyword \texttt{album}. After retrieving this memory, the system changes how it selects evidence paragraphs and limits the number retained in the context. Under this limit, the selected context omits \texttt{Winner (band)}, which contains the evidence needed to identify who formed \texttt{WINNER}.

\Needspace{8\baselineskip}
\begin{caseListing}{Candidate response to the test question}
{
  "final_answer": "YG Entertainment",
  "supporting_facts": [
    {"title": "2014 S/S", "sentence_id": 1}
  ],
  "atomic_facts": [
    {"subject": "2014 S/S", "relation": "is debut album of",
     "object": "WINNER",
     "title": "2014 S/S", "sentence_id": 0},
    {"subject": "WINNER", "relation": "formed by",
     "object": "YG Entertainment",
     "title": "2014 S/S", "sentence_id": 1}
  ],
  "reasoning": {"operation": "bridge", "fact_indices": [0, 1],
                "result": "YG Entertainment"}
}
\end{caseListing}

The answer remains \texttt{YG Entertainment}, but the response now attributes the \texttt{formed by} relation to sentence 1 of \texttt{2014 S/S}. Its submitted supporting fact list contains only that sentence and omits both gold supporting facts.

\begin{caseListing}{Sentence 1 of 2014 S/S}
It was released on August 12, 2014 by the group's record label, YG Entertainment.
\end{caseListing}

This sentence concerns the album's release by a record label. It does not establish who formed the group. The answer happens to remain correct, so Answer F1 remains 1, while supporting fact F1 and Joint F1 fall to 0. On the full test set, the accepted update lowers the mean Joint F1 from 54.55\% to 48.00\%: 13 items decline and none improve. This example shows how a memory can repair a repeated validation question while causing evidence selection to lose a required source on a new question.

\endgroup

\FloatBarrier

\end{document}